\documentclass{egpubl}
\usepackage{STAG2025}
 
\WsPaper           

\usepackage[T1]{fontenc}
\usepackage{dfadobe}  

\biberVersion
\BibtexOrBiblatex

\electronicVersion
\PrintedOrElectronic

\ifpdf \usepackage[pdftex]{graphicx} \pdfcompresslevel=9
\else \usepackage[dvips]{graphicx} \fi

\usepackage{egweblnk} 

\usepackage{graphicx}

\newcommand{\etal}{\textit{et al.}}
\usepackage{amsmath}
\usepackage{booktabs}

\title{Learning to Predict Aboveground Biomass from RGB Images with 3D Synthetic Scenes}

\author[S. Zuffi]
{\parbox{\textwidth}{\centering Silvia Zuffi
\\
{\centering Institute for Applied Mathematics "Enrico Magenes", CNR-IMATI, Milan, Italy}
}
}

\begin{document}
\teaser{
 \includegraphics[width=1.0\linewidth]{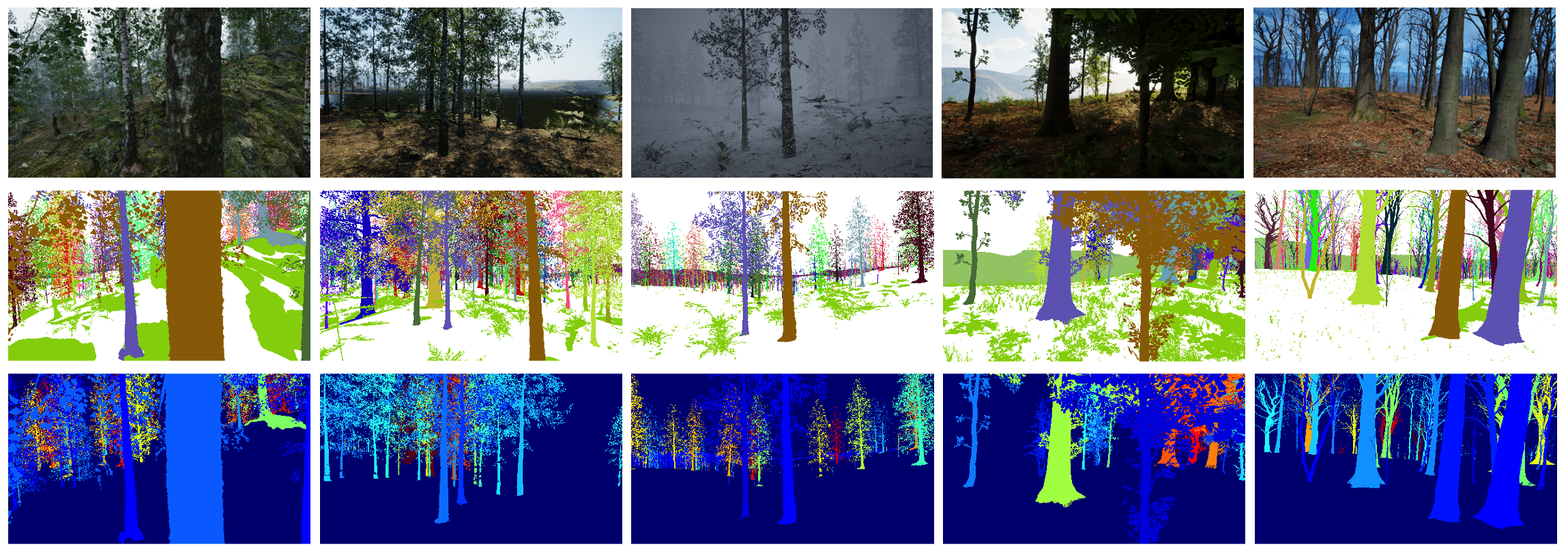}
 \centering
  \caption{We estimate Aboveground Biomass (AGB, $kg/m^2$) from RGB images using a synthetic 3D dataset. The task is formulated as a regression problem that predicts AGB density maps from RGB inputs. An AGB density map represents the per-pixel density of the AGB in the scene; integrating the density map over the image yields the AGB estimate.
 To train our regression network, we use RGB renderings generated from a synthetic 3D dataset of arboreal scenes, which provides instance segmentations and tree geometric attributes for computing corresponding ground-truth density maps.
 Top: RGB image; Middle: instance segmentation; Bottom: AGB density map. 
 Each AGB density map is a single-channel 2D image, 
 where a pixel stores the biomass of the corresponding tree divided by the plot area and image area in pixels. Large trees or trees that are far away, and thus are small objects in the image, have larger values. We visualize the AGB map with a color coding aligned with the visible spectrum (small values are blue, then green, yellow and large values in red). Note that we apply a gamma correction to the density map for better visualization.}
\label{fig:teaser}
}
\maketitle 
\begin{abstract}
Forests play a critical role in global ecosystems by supporting biodiversity and mitigating climate change via carbon sequestration. Accurate aboveground biomass (AGB) estimation is essential for assessing carbon storage and wildfire fuel loads, yet traditional methods rely on labor-intensive field measurements or remote sensing approaches with significant limitations in dense vegetation.
In this work, we propose a novel learning-based method for estimating AGB from a single ground-based RGB image. We frame this as a dense prediction task, introducing AGB density maps, where each pixel represents tree biomass normalized by the plot area and each tree’s image area. 
We leverage the recently introduced synthetic 3D SPREAD dataset, which provides realistic forest scenes with per-image tree attributes (height, trunk and canopy diameter) and instance segmentation masks. Using these assets, we compute AGB via allometric equations and train a model to predict AGB density maps, integrating them to recover the AGB estimate for the captured scene. 
Our approach achieves a median AGB estimation error of $1.22\,kg/m^2$ on held-out SPREAD data and $1.94\,kg/m^2$ on a real-image dataset. To our knowledge, this is the first method to estimate aboveground biomass directly from a single RGB image, opening up the possibility for a scalable, interpretable, and cost-effective solution for forest monitoring, while also enabling broader participation through citizen science initiatives.
\end{abstract}    
\section{Introduction}
\label{sec:intro}
Forests are a vital component of the Earth's global ecosystem, serving as habitats for countless species while providing food and oxygen. Most importantly, they play a crucial role in mitigating climate change by capturing and storing carbon dioxide \cite{Pan2013_AREES}. Despite their relevance, forests face significant threats from human exploitation and wildfires, and it is of high importance to monitor their status and evolution. Among many factors, the accurate estimate of forest biomass offers valuable insight into carbon storage capacity and wildfire fuel loads, both critical factors in effective forest management.

Biomass refers to the total mass of living organisms within a given area, typically expressed as weight per unit area. In forests, this includes trees, shrubs, understory plants, and roots~\cite{Tadese2019AGBEstimation}. Many studies specifically focus on biomass excluding roots, and use the term Aboveground Biomass (AGB) to indicate the material density outside the ground. The AGB estimation from single RGB images is the focus of our study. 

Traditional methods for aboveground biomass estimation are based on direct measurements of tree height, trunk and canopy diameter. Then, such measurements are used in simple allometric equations that allow computing biomass from measurements like tree height and DBH (diameter at breast height). While this approach is accurate, obtaining such measurements on the field is time-consuming, often relies on destructive field harvesting, and it is thus not scalable ~\cite{Chave2014ImprovedAM,Yang2022_FE}.

Alternative methods for AGB estimation exploit remote sensing observations in combination with ground-level surveys~\cite{SU2025103002}, or explicitly apply image analysis methods for the estimation of tree counts and height~\cite{Liu2024Improving}.
Studies have found that tree height is the most important variable that correlates with biomass~\cite{Chave2014ImprovedAM}. While in the past this information was difficult to obtain, with recent methods based on aerial and satellite observation, tree canopy height can be estimated over large regions, providing essential information to define AGB maps over the entire surface of the planet~\cite{Lang2023,Sialelli2024}.
Such methods are however limited from being based only on observations from the above, require hyperspectral data and/or specialized equipment, and can be inaccurate in cases of dense stratified forests. 

In this work, we address the problem of automatically estimating forest aboveground biomass from ground-based RGB images. Estimating AGB from these images has the potential to provide a high spatial resolution assessment of forest biomass, as well as providing insights into forest structure and composition. In addition, being based only on color images, our method can represent a useful tool for citizen science initiatives.

One approach for computing biomass from a single image could involve detecting trees and estimating their dimensions, followed by calculating biomass from these measurements. However, this approach would require a calibrated system capable of providing metric measurements and would be limited to only the visible parts of the trees in the image, which is often a partial observation in case of large individuals. 
Light detection and ranging (LiDAR) sensors have been proposed to capture forests metrics. While accurate, these solutions require specialized equipment~\cite{Demol2022Estimating,howie2024}.
Another possible approach is to leverage current image-to-3D reconstruction methods~\cite{Li2024SVDTree, Lee2025TreeDFusion, Li2021} to compute the volume of trees and thus their biomass. Unfortunately, current techniques only work satisfactorily for single, independent trees, while in this work we aim at estimating biomass for dense forests.

\begin{figure*}[t]
    \centering
    \includegraphics[width=\linewidth]{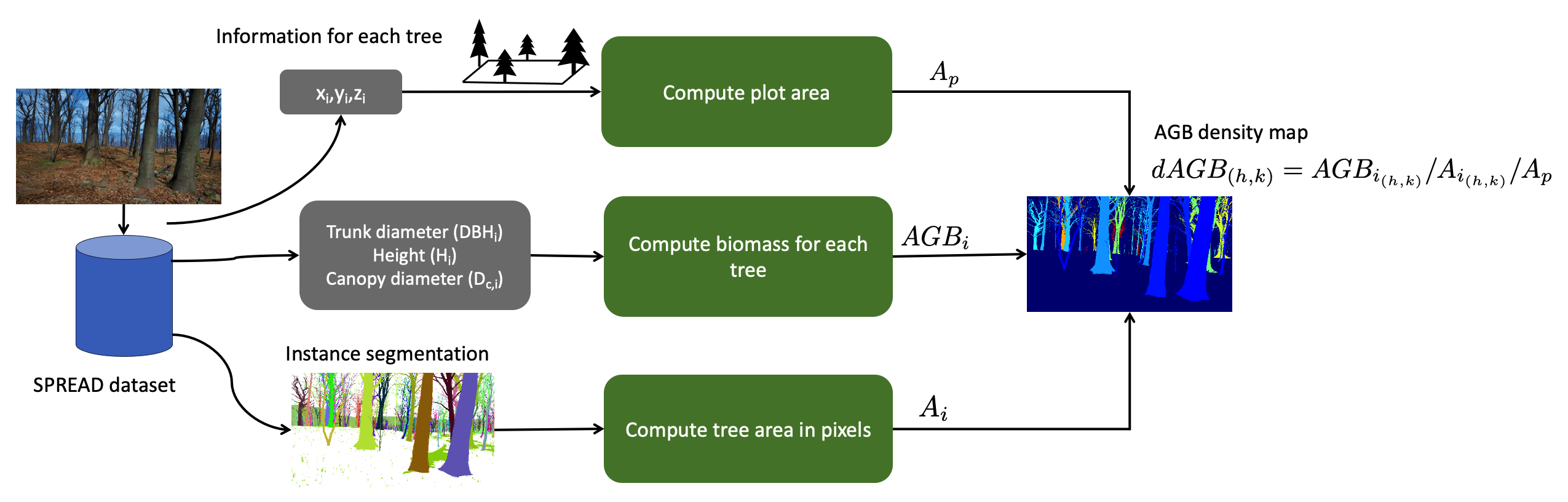}
     \caption{Pipeline to generate the AGB density maps. The density map $dAGB$ at pixel location $(h,k)$ stores the biomass of the tree at that location $AGB_{i_{(h,k)}}$ divided by tree area in pixels $A_i$ and plot area $A_p$.}
    \label{fig:agb_pipeline}
\end{figure*}

We aim to develop an approach that does not require a calibrated system and remains robust to camera movements. To achieve this, we adopt a learning-based methodology inspired by modern image processing architectures, which have demonstrated strong performance in tasks such as image segmentation and depth estimation.
Given these advancements, we formulate aboveground biomass estimation from a single image as a dense prediction task. To accomplish this, we introduce a novel representation, the \emph{AGB density map} (see Fig.~\ref{fig:teaser}). 
An AGB density map is a single-channel 2D image aligned with the input RGB image, where each pixel represents the biomass of the corresponding tree divided by the plot area and the tree’s image area in pixels. This formulation ensures that the integral over the entire AGB density map yields the AGB of the plot area ($kg/m^2$) captured by the camera (See Fig.~\ref{fig:agb_pipeline}). In principle, this representation is invariant to small camera movements, as long as trees do not leave or enter the captured scene.

\begin{figure*}[t]
    \centering
    \includegraphics[width=0.97\linewidth]{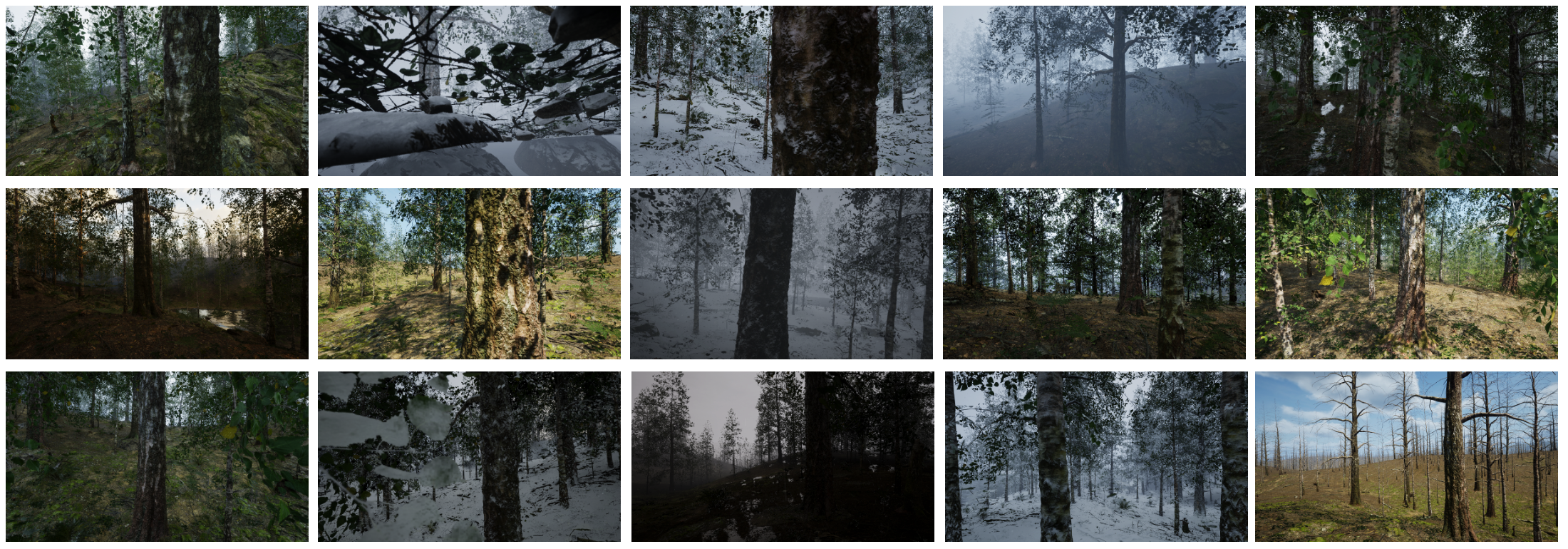}
     \caption{Examples from the SPREAD dataset~\cite{feng2024spread}, Birch Forest.}
    \label{fig:spread_bi}
\end{figure*}

\begin{figure*}[t]
    \centering
    \includegraphics[width=0.97\linewidth]{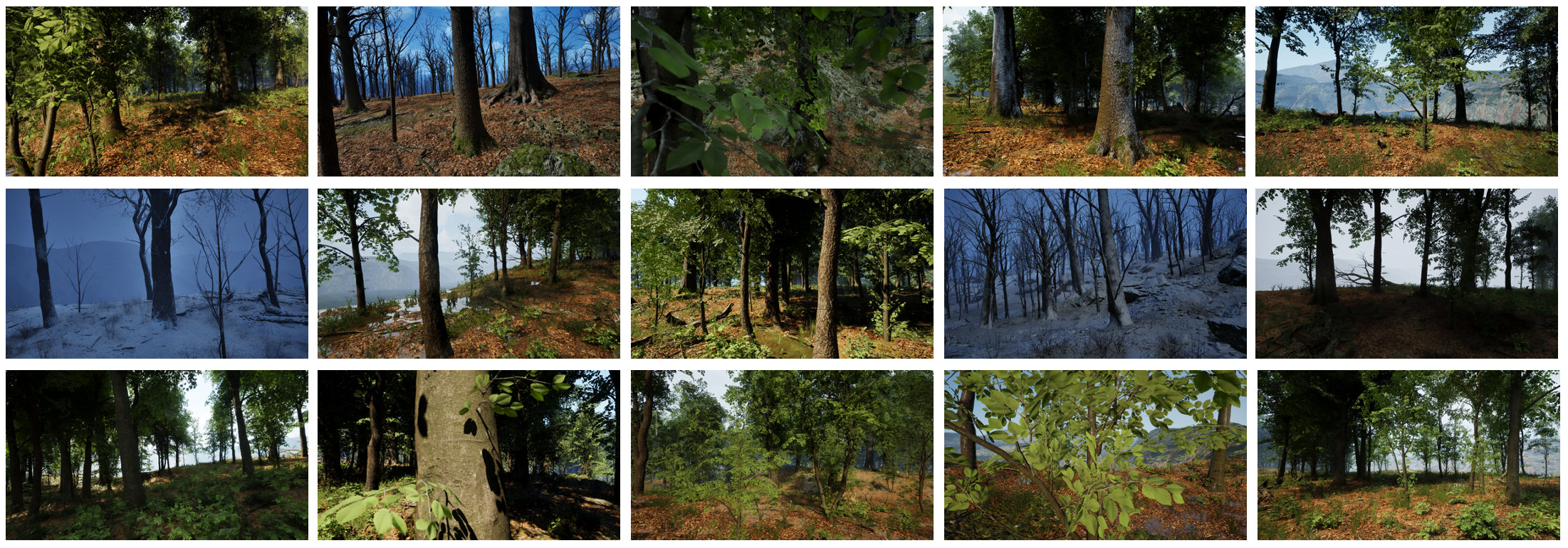}
     \caption{Examples from the SPREAD dataset~\cite{feng2024spread}, Broadleaf Forest.}
    \label{fig:spread_bl}
\end{figure*}

We define AGB density maps by leveraging a recently introduced synthetic 3D dataset, SPREAD~\cite{feng2024spread}, created with Unreal Engine 5, that includes highly realistic forest scenes (see Fig.~\ref{fig:spread_bi} and Fig.~\ref{fig:spread_bl}), along with per-image assets information of tree height, DBH, canopy diameter, and location in space. Instance segmentation and depth maps are also defined. 
We exploit the tree attributes to compute the biomass of each tree using allometric equations, while the instance segmentation masks are used to compute the area of each tree in pixels. Using the tree coordinates in 3D space, we compute the plot area as the ground-level bounding box encompassing all trees. Based on this information, we define the AGB density map. 

We train a model in a supervised manner to regress AGB density maps from RGB images, and subsequently integrate these maps to obtain the total AGB estimate. One might note that AGB could be predicted directly from the images by regressing to a single scalar value. While this is indeed possible, we argue that predicting the AGB density map as an intermediate representation yields a more interpretable system, as the map reveals which image regions are associated with higher biomass. We further justify this design choice in the Experiments section.

We implement the regressor as a transformer-based architecture, using as a backbone a Swin Transformer~\cite{Liu2021_swint}.
We perform a quantitative evaluation on a held-out set from the SPREAD dataset, achieving AGB estimates with a median absolute error of $1.22\,kg/m^2$. 
Although we train exclusively on synthetic images, we argue that if the images generated using a high-realism game engine are indistinguishable from real images, the learned model should perform equally well on real-world data.
We present, on real images, qualitative results by regressing the AGB of a set of images and ordering the images by increasing AGB. A visual inspection confirms that the AGB estimate is consistent with the density and size of trees in the images. 

To the best of our knowledge, no real-world dataset currently exists for evaluating AGB estimates derived from RGB images of real scenes, which limits the possibility of performing a quantitative assessment on real data.
To partially address this limitation, we assembled a dataset of real images with ground-truth AGB values by leveraging data from the Italian National Forest Inventory \cite{INFC}, which provides AGB measurements, vegetation types, and geographic coordinates. Using these coordinates, we collected corresponding Google Street View images where available near the inventory locations.
On this dataset, our model achieves a median absolute error of $1.94\,kg/m^2$.

One might argue that the problem we address inherently involves a spatial component requiring metric estimation, and that in our quantitative experiments on held-out SPREAD data,
we circumvent the monocular scale ambiguity by training and testing on images that have been generated with similar cameras.
However, because the synthetic dataset can be customized to specific camera models (e.g., smartphone cameras), we envision applications where the system is trained and optimized for a particular imaging device. This is feasible, as synthetic images can be generated with arbitrary camera parameters.

To the best of our knowledge, our approach represents the first attempt to directly estimate aboveground biomass from a single RGB image, opening new opportunities for scalable and accessible forest monitoring, also enabling broader participation through citizen science initiatives.

\section{Related Work}
\label{sec:previous}
\textbf{Aboveground Biomass Estimation}
Traditional approaches to AGB estimation involve direct field measurements, which can be categorized into destructive and non-destructive methods. Destructive methods involve harvesting and weighing vegetation to obtain precise biomass data. This is a very time-consuming and labor-intensive solution, difficult to apply at large scale. Not to mention that the approach requires tree harvesting, which could be prohibited in natural reserves~\cite{Ma2024DevelopmentFA}.
Non-destructive methods involve identifying a set of sample plots within the survey area, and measuring variables like tree height and diameter at breast height (DBH) for all the trees in the plot. Then, given the species-specific wood density, the biomass of each tree is calculated using allometric equations. This solution requires access to the trees for measurement. 
A widely used approach is to derive regression methods from remote sensing indices to field AGB estimates. Such methods have reported errors up to $50\%$ \cite{Yang2022_FE}. 
In recent years, LiDAR technology has been used in estimating above-ground biomass because of its ability to capture detailed three-dimensional structural information of vegetation. Airborne LiDAR systems, in particular, provide high-resolution data that enable accurate assessment of forest biomass across various ecosystems~\cite{Chan2021}. 
Recent advancements also include the fusion of LiDAR with other remote sensing data. 
Specifically, Lang \etal ~\cite{Lang2023} fuse sparse height data from NASA's Global Ecosystem Dynamics Investigation (GEDI) space-borne LiDAR mission with dense optical satellite images from Sentinel-2 and present the first global canopy height map at a 10-meter ground sampling distance. Since tree height highly correlates with biomass, accurate canopy height maps are an important analysis tool to assess biomass, especially if integrated with field observations.
A recent study evaluates the accuracy of GEDI data in Japanese coniferous forests by comparing with airborne LiDAR data for GEDI-derived terrain elevations, canopy heights, and aboveground biomass density (AGBD), reporting for the latter an average rRMSE of $52.79\%$~\cite{LI2024100144}.
A comprehensive dataset that integrates AGB reference data from the GEDI mission with imagery from Sentinel-2 and PALSAR-2 satellites is introduced in~\cite{Sialelli2024}.
 
\noindent\textbf{Density Prediction from RGB Images}
The estimation of AGB maps from RGB images is a dense regression task, which shares similarities with depth estimation and semantic image segmentation. 
A common trend in dense prediction tasks is to take advantage of the visual knowledge embedded in image models trained on large image datasets.
A recent approach, Marigold~\cite{marigoldke2023}, leverages pre-trained diffusion-based image generation models, specifically Stable Diffusion~\cite{rombach2022high}, to address the task of monocular depth estimation. The method fine-tunes the denoising UNet~\cite{unetronneberger2015} of a pre-trained Latent Diffusion Model (Stable Diffusion v2) on $74K$ synthetic data samples. Despite not being trained on real data, the model exhibits zero-shot generalization to real scenes of high complexity. Recently, the model has been extended to video~\cite{ke2024rollingdepth}.   
Depth Anything~\cite{depth_anything_v1} relies on DINOv2~\cite{oquab2023dinov2}, a foundational model trained self-supervised on 142M images. DINO models~\cite{caron2021emerging} can be fine-tuned by adding learnable adaptation layers in the popular LoRA framework~\cite{hu2022lora}.
 
In terms of architectures, Vision Transformers (ViTs) are an effective alternative to convolutional networks for dense prediction tasks~\cite{Ranftl2021}. 
Among them, Swin Transformers (Shifted Window Transformers) are a type of vision transformer designed for efficient and scalable image processing, proposed to serve as a versatile backbone for various computer vision tasks~\cite{Liu2021_swint}. Unlike standard Vision Transformers, which operate on fixed-size image patches, Swin Transformers use a hierarchical structure and shifted window attention to model long-range dependencies and visual data at different scales, while maintaining computational efficiency. Swin Transformers have proven to be a good choice for various computer vision tasks, including image classification, object detection and semantic segmentation~\cite{drapier2025combiningtransformerscnnsefficient, Liu2021_swint}. 

\section{Method}
\label{sec:method}
\subsection{Preliminaries}
\textbf{Allometric Equations}
Allometric models allow inferring the aboveground biomass of oven-dry trees from census data. When trunk diameter, total tree height and wood specific gravity are available, in a study using a globally distributed database of tropical forests, subtropical forests, and woodland savannas, a single model was found to be suitable for modeling the different vegetation types~\cite{Chave2014ImprovedAM}. Such model takes the form:
\begin{equation}
    AGB_i = 0.0673 \times (\rho\,DBH^2H)^{0.967},
\label{eq:agb}
\end{equation}
where DBH is the tree diameter at breast height expressed in $cm$, H is the tree height expressed in $m$, and $\rho$ is the wood specific density defined as oven-dry wood mass divided by its green volume, and expressed in $g/cm^{-3}$. While designed with tropical forest data, the model is widely adopted as it performs well across forest types and bioclimatic conditions. Note that in the literature, the term AGB is used to refer to both the biomass per land area and the biomass of a single tree. We adopt here this common notation; to avoid confusion we indicate the AGB for a single tree,  expressed in $kg$, with a subscript $i$. 

\noindent\textbf{SPREAD Dataset} SPREAD ~\cite{feng2024spread} is a recently introduced 3D synthetic dataset designed for image-based visual tasks on arboreal landscapes. 
The dataset provides RGB and depth images, semantic and instance segmentation maps, and tree parameters of DBH, height, and canopy diameter. Additionally, the metadata includes the spatial location of each tree.
SPREAD consists of 13 photo-realistic 3D virtual scenes representing diverse environments, from forests to urban areas and plantations. Being generated with Unreal Engine 5, the dataset is highly realistic, with visual variation achieved also by rendering the synthetic scenes under different weather conditions (Fig.~\ref{fig:spread_bi} and Fig.~\ref{fig:spread_bl}). 

\subsection{Aboveground Biomass Density Maps}
To define the AGB density maps, we first retrieve the list of all trees in the scene. 
Then, for each image, we load the instance segmentation map. Each color in the map corresponds to an index, and each index corresponds to a tree name in the global list of trees. We convert the instance segmentation map into a 1-channel map with indices, and then use this map to compute the pixel area for each tree. 
We compute the biomass for each tree with Equation~\ref{eq:agb}.
We use the tree coordinates to determine the plot area $A_p$, defined as the ground-level bounding box. Specifically, $A_p$
 is computed as the area of the rectangle formed by the minimum and maximum ground coordinates of all trees in the scene.
We then generate the AGB density map by assigning to each pixel the value of the corresponding tree biomass divided by its image area (in pixels) and plot area (in $m^2$). Note that we discard the trees that have a pixel area smaller than $2\%$ of the image. With this representation, we compute the total AGB for an image as the integral of the density map:
\begin{equation}
    AGB = \sum_{i} \frac{AGB_i}{A_p} = \sum_{h,k}dAGB[h,k],
\label{eq:dagb}
\end{equation}
where $AGB_i$ is the biomass of the $i^{th}$ tree, $A_p$ is the plot area, and $dAGB[h,k]$ is the value of the AGB density map (see Fig.~\ref{fig:agb_pipeline}) at the pixel location $(h,k)$. 

Note that we assume that the forest is dense and measure the plot area using the bounding box defined by the visible trees. When this assumption is violated, the density map value can become very large, and we filter such values when creating the training set. By computing the bounding box, we make the implicit assumption that the ground is flat. A more accurate plot area measurement to be used in mountain scenery would take into consideration the base coordinates of all the trees to estimate an uneven plot area. 

\subsection{AGB Prediction}
\begin{figure*}[t]
    \centering
    \includegraphics[width=\linewidth]{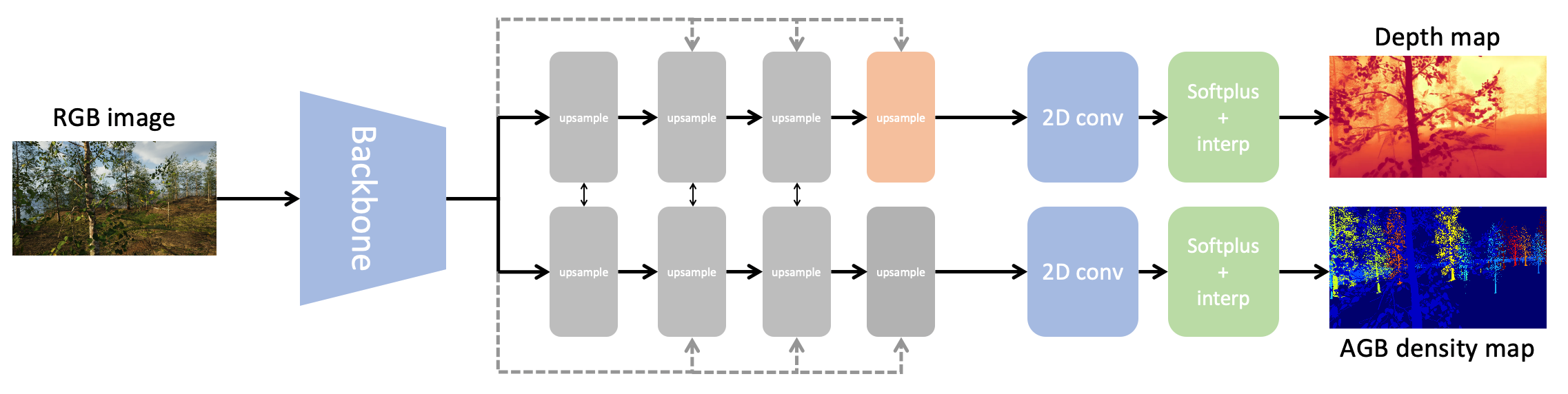}
     \caption{Network Architecture. Our model is composed by a Swin Transformer backbone, and a set of blocks with skip connections. A 2D convolution block follows, with then a SoftPlus function and bilinear interpolation. The image features pass through two paths, one to generate the AGB density map, and one to generate the depth map. The first three upsampling blocks are shared.}
    \label{fig:network}
\end{figure*}

We design a regression network to predict aboveground biomass density maps from an RGB image. Our model is based on a Swin Transformer backbone and a set of upsampling blocks, followed by 2D convolution, SoftPlus function and bilinear interpolation (Fig.~\ref{fig:network}). We consider Swin Transformers as, compared with ViTs, they are more suitable for dense prediction tasks~\cite{Liu2021_swint}. 
The input image is preprocessed and resized to 
(224,224) pixels before being passed through the backbone network to obtain multi-scale features. 
A set of $4$ decoders, each consisting of a convolutional layer, batch normalization, and a non-linear activation function, progressively upsamples the features while incorporating skip connections. The upsampled features are then processed through a convolutional layer and a SoftPlus function, and finally interpolated to match the output size. 
During training, backbone features optionally follow a parallel path for depth map prediction. 
We found that, when incorporating depth estimation, the best option is to have the first $3$ upsampling blocks shared with the RGB path, while the last block is specific for depth prediction. A 2D convolutional layer and a SoftPlus function complete the depth prediction path. 
We train the network to minimize the following loss:

\begin{equation}
    \begin{split}
        Loss &= L1(dAGB, dAGB_{GT})  \\
             &\quad + \alpha_1\, \left| AGB, AGB_{GT}\right|   \\
             &\quad + \alpha_2\,L1(D, D_{GT}),
    \end{split}
    \label{eq:loss}
\end{equation}

where $dAGB$ is the predicted AGB density map, $AGB$ is the total AGB in the image, computed as the integral of the density map. 
We found that including this total-AGB loss term improves prediction accuracy.
 D is the normalized depth map, and the subscript $GT$ indicates the ground-truth training data. We set $\alpha_1=1e^{-5}$, which corresponds to the image area, and $\alpha_2=1$.

\begin{figure*}[t]
\centering
\centering
\includegraphics[width=0.33\linewidth]{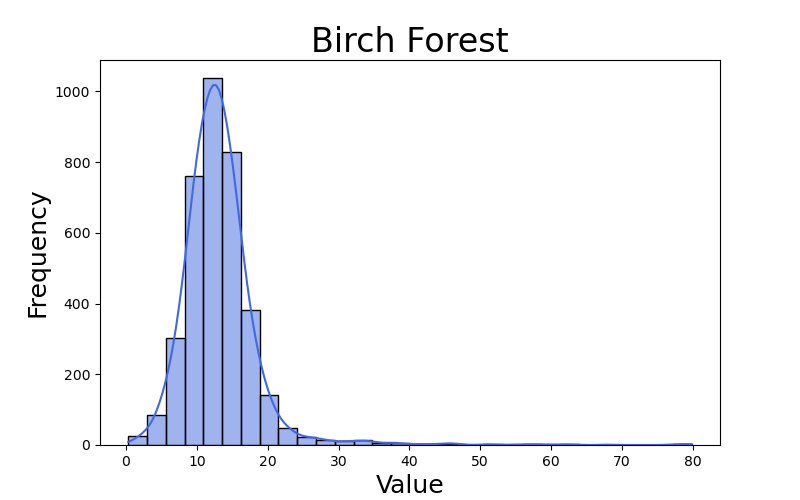}
\hspace{-0.3cm}
\includegraphics[width=0.33\linewidth]{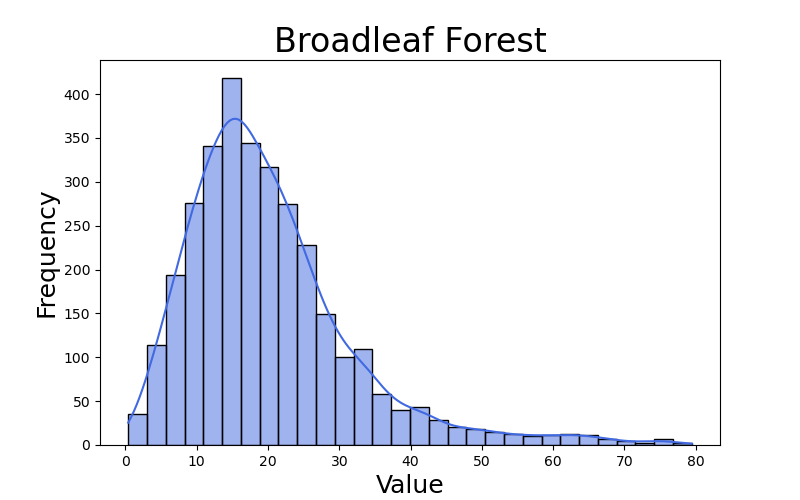}
\hspace{-0.3cm}
\includegraphics[width=0.33\linewidth]{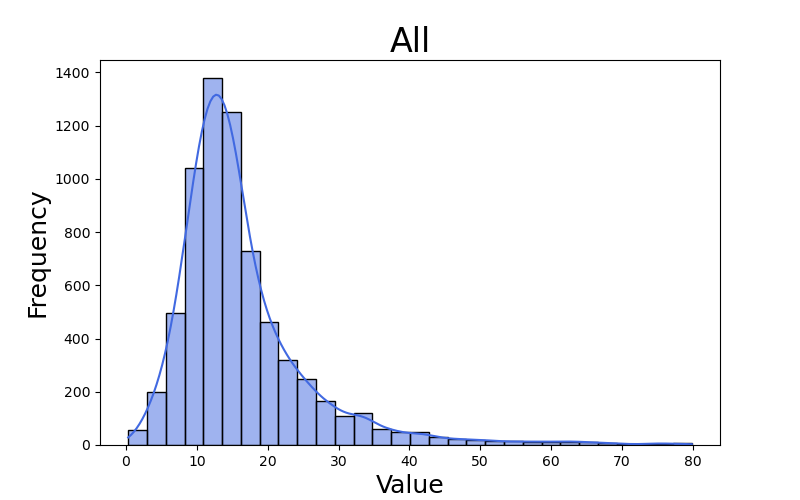}\\
\hspace{-0.3cm}
\caption{
We show the histograms of the total AGB values in the per-scene datasets and in the total dataset. 
}
\label{fig:histograms}
\end{figure*}

\section{Experiments}
\begin{figure*}
    \centering
    \includegraphics[width=0.28\linewidth]{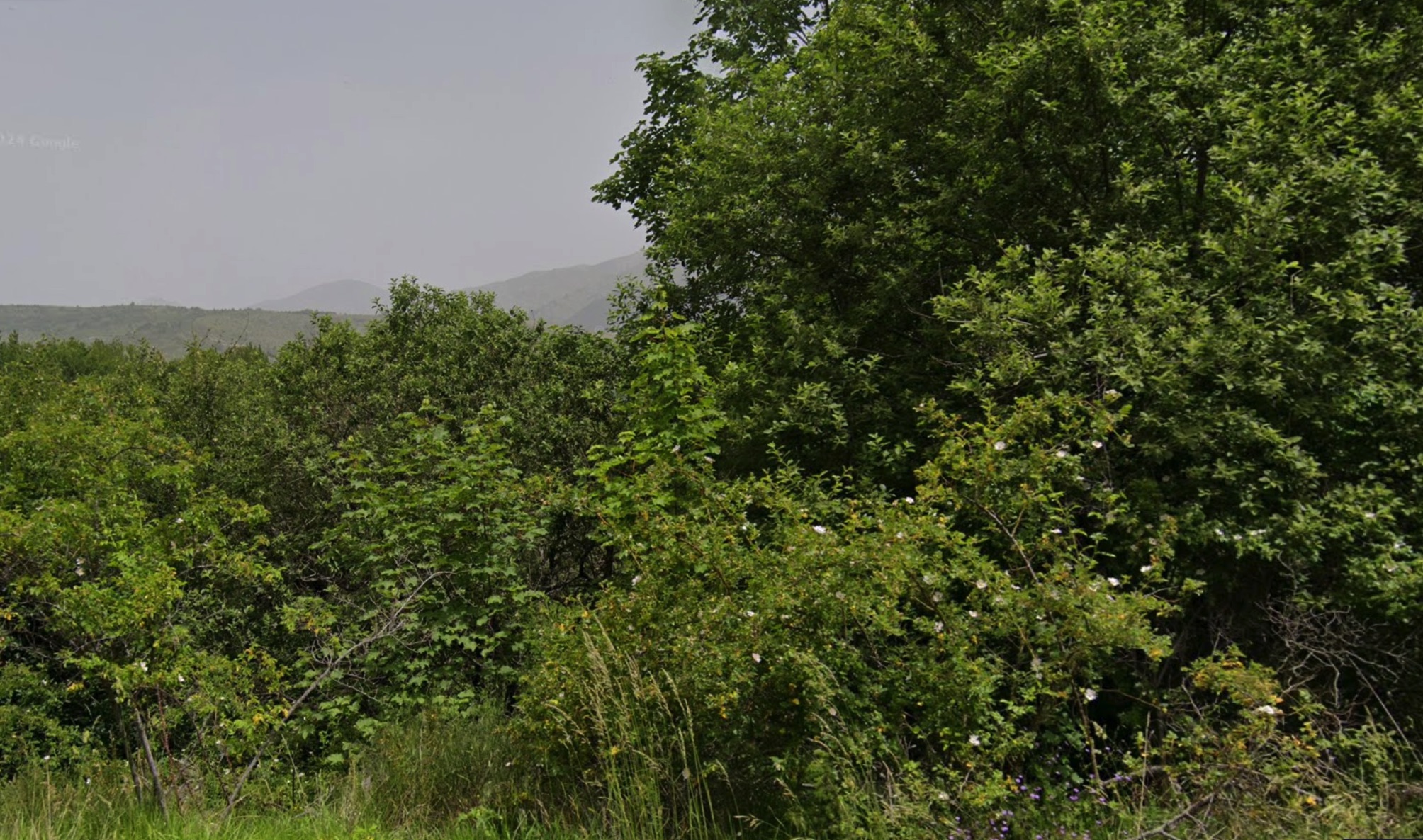}
    \includegraphics[width=0.28\linewidth]{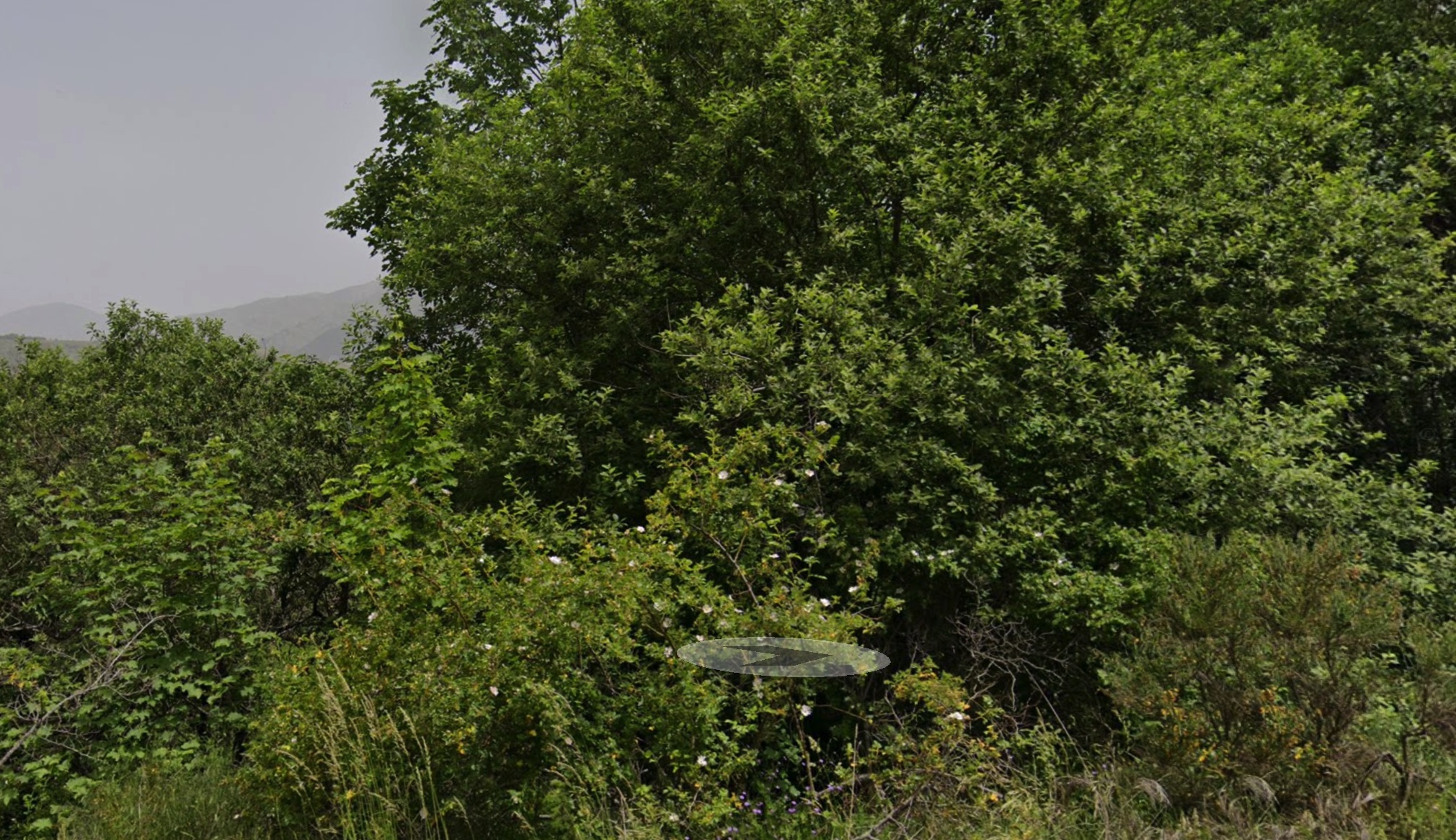}
    \includegraphics[width=0.28\linewidth]{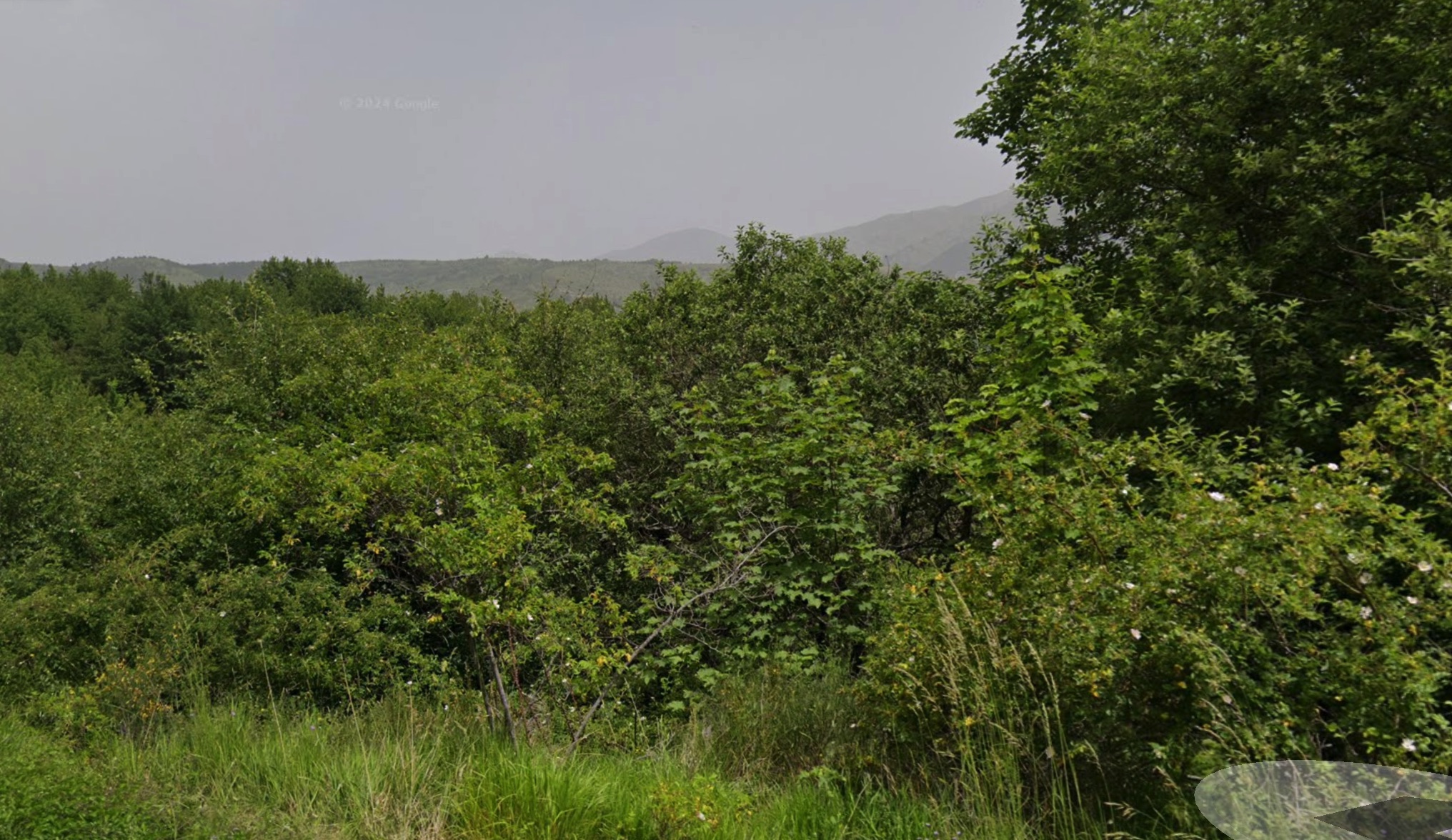}
     \centering
   \includegraphics[width=0.28\linewidth]{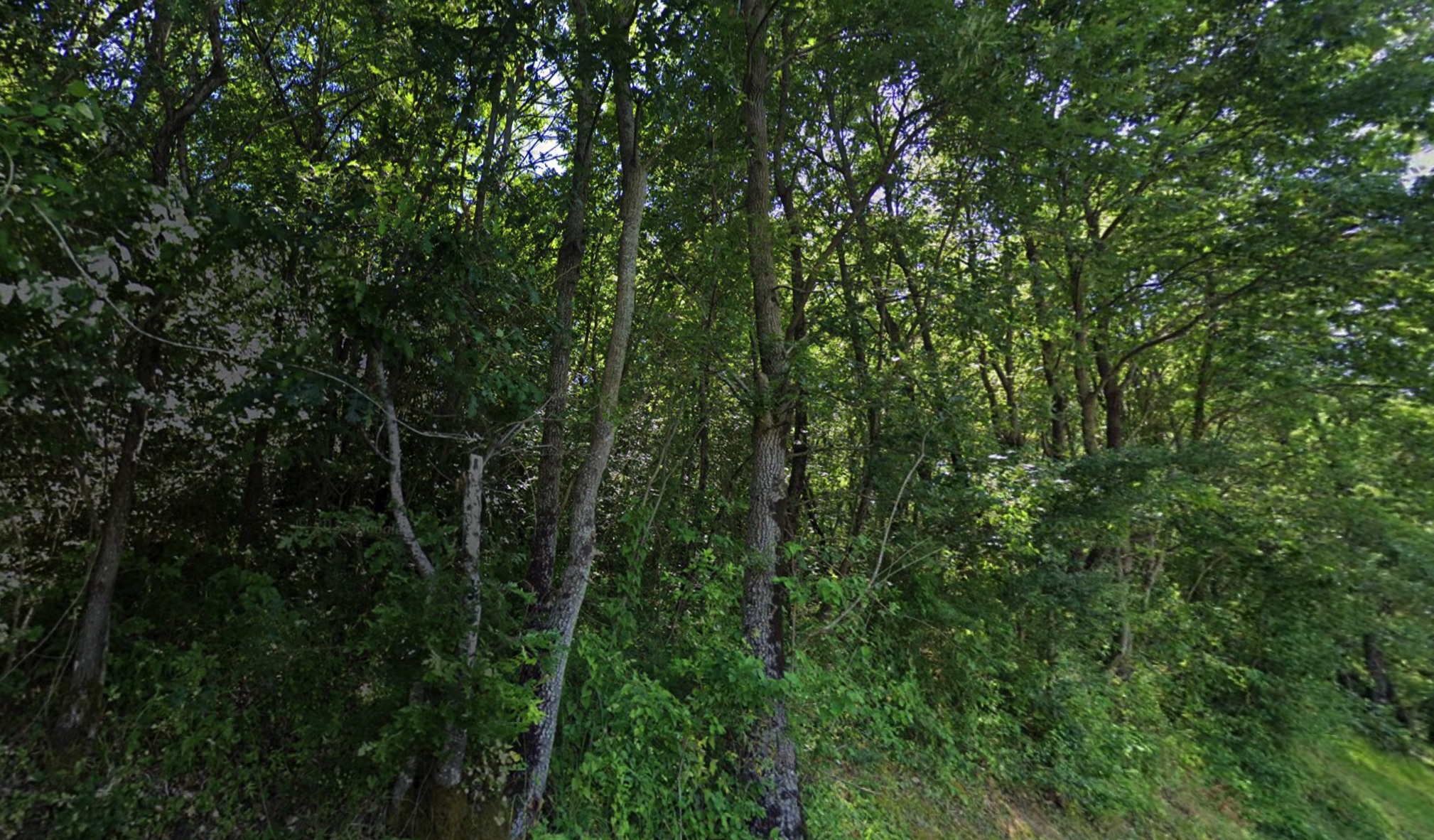}
   \includegraphics[width=0.28\linewidth]{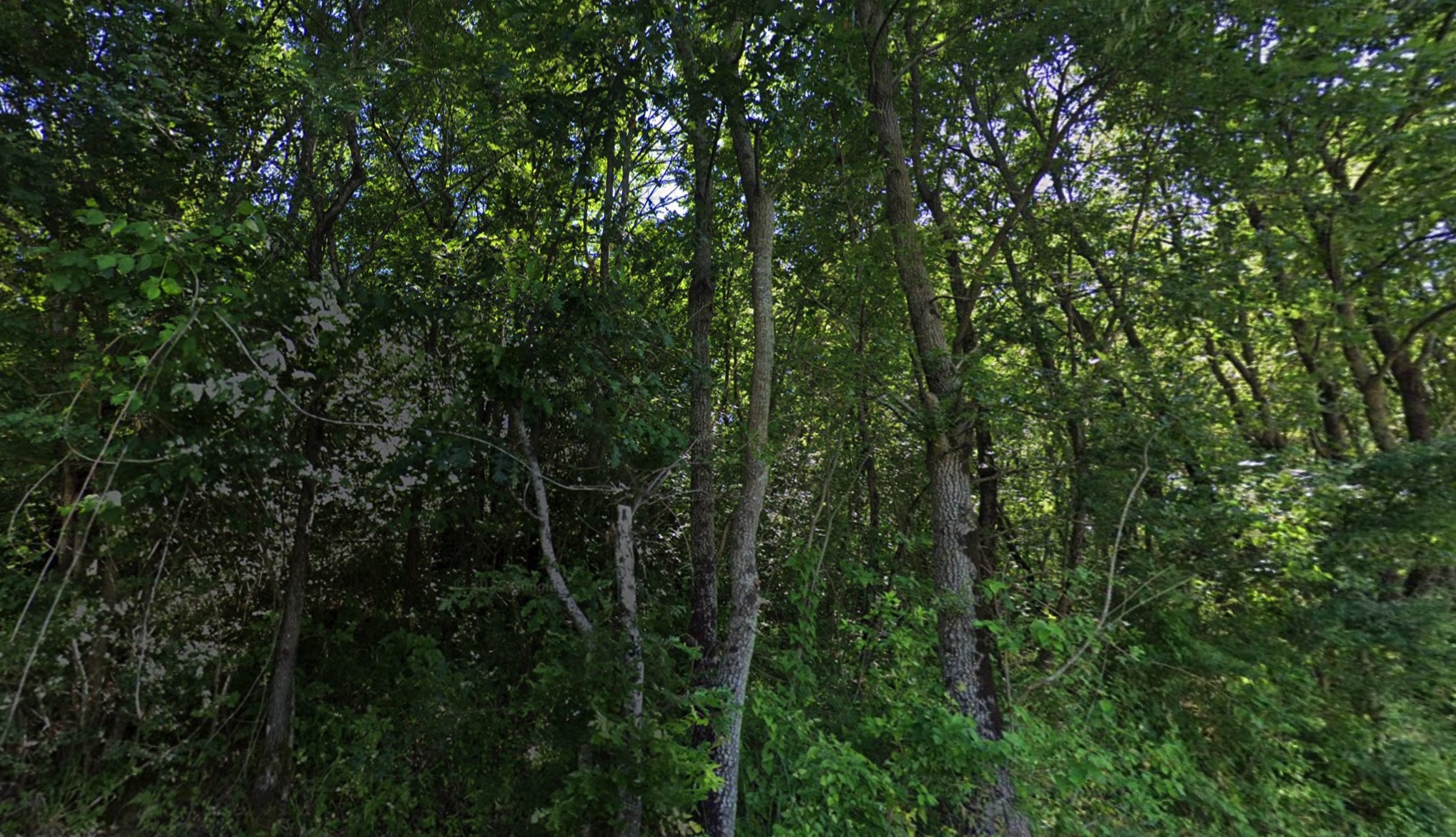}
   \includegraphics[width=0.28\linewidth]{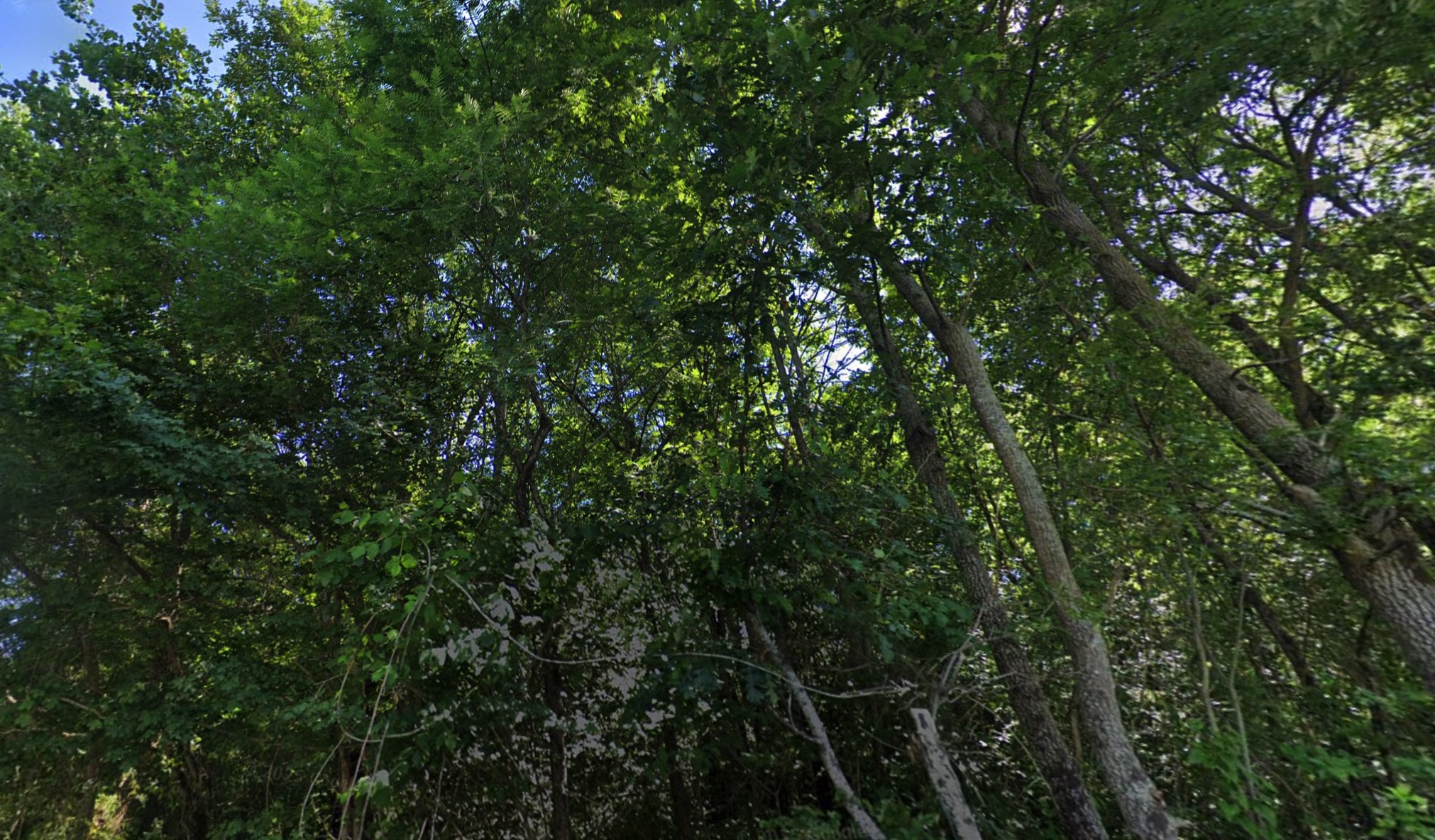}
    \centering
    \includegraphics[width=0.28\linewidth]{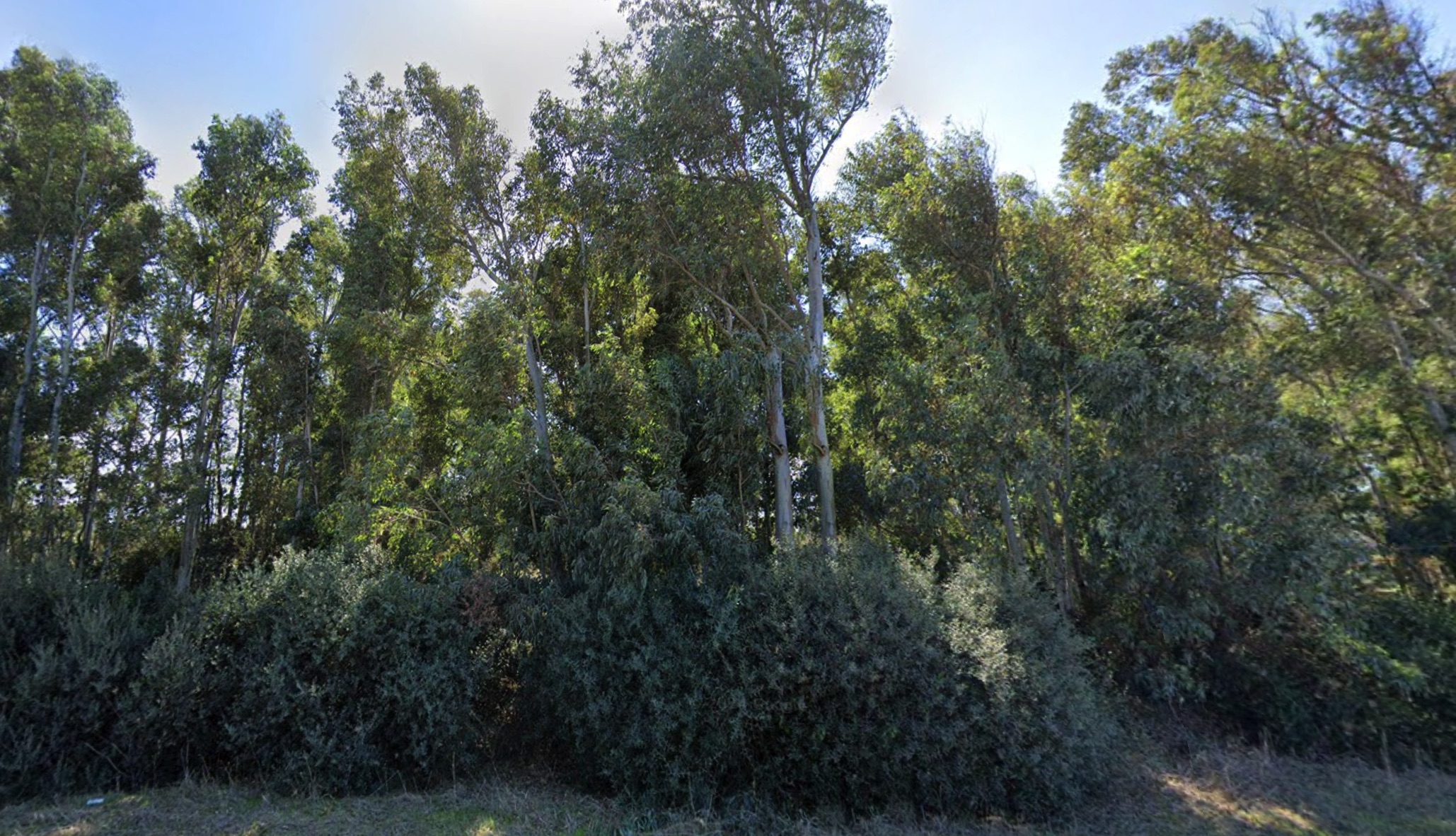}
    \includegraphics[width=0.28\linewidth]{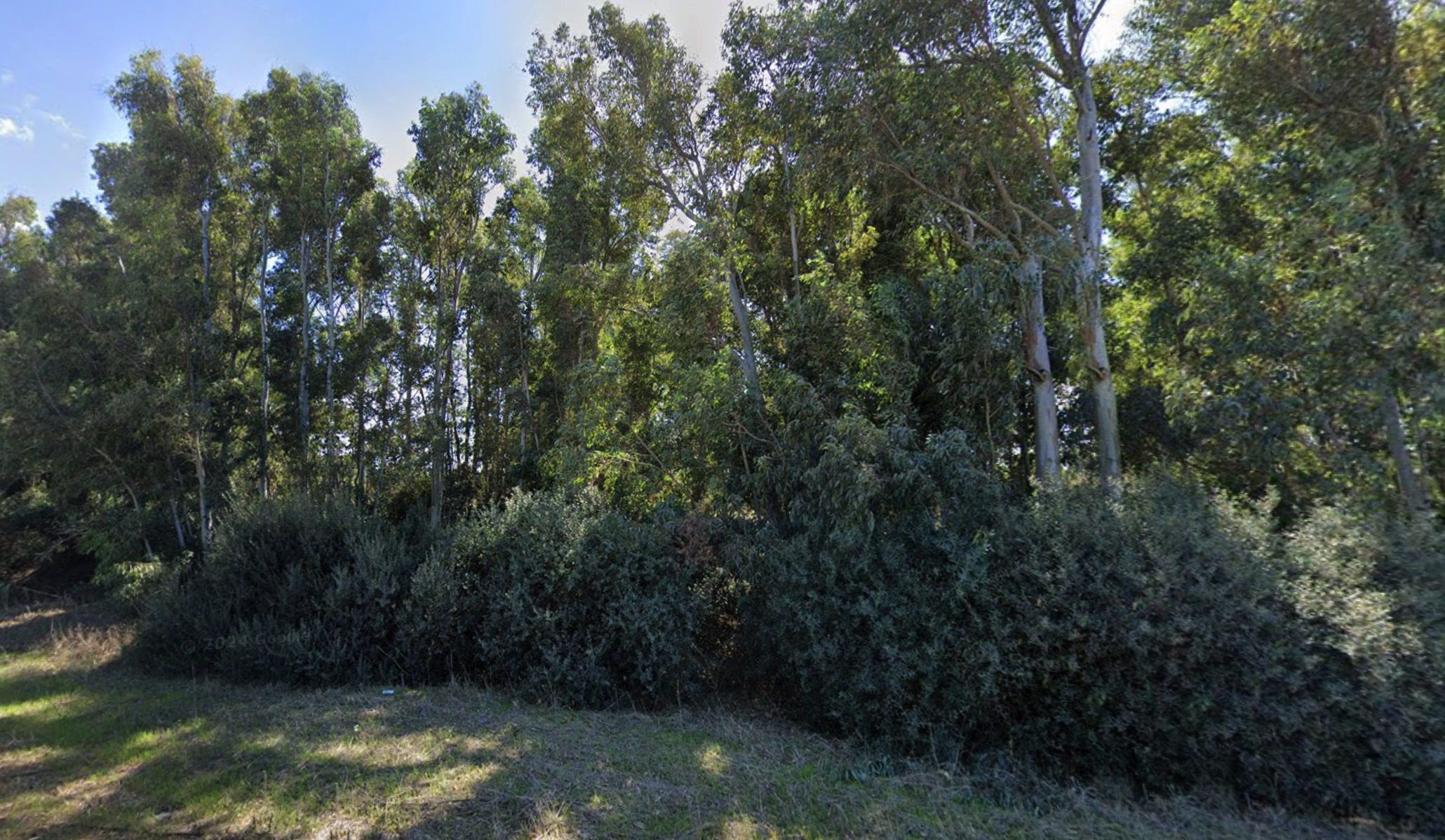}
    \includegraphics[width=0.28\linewidth]{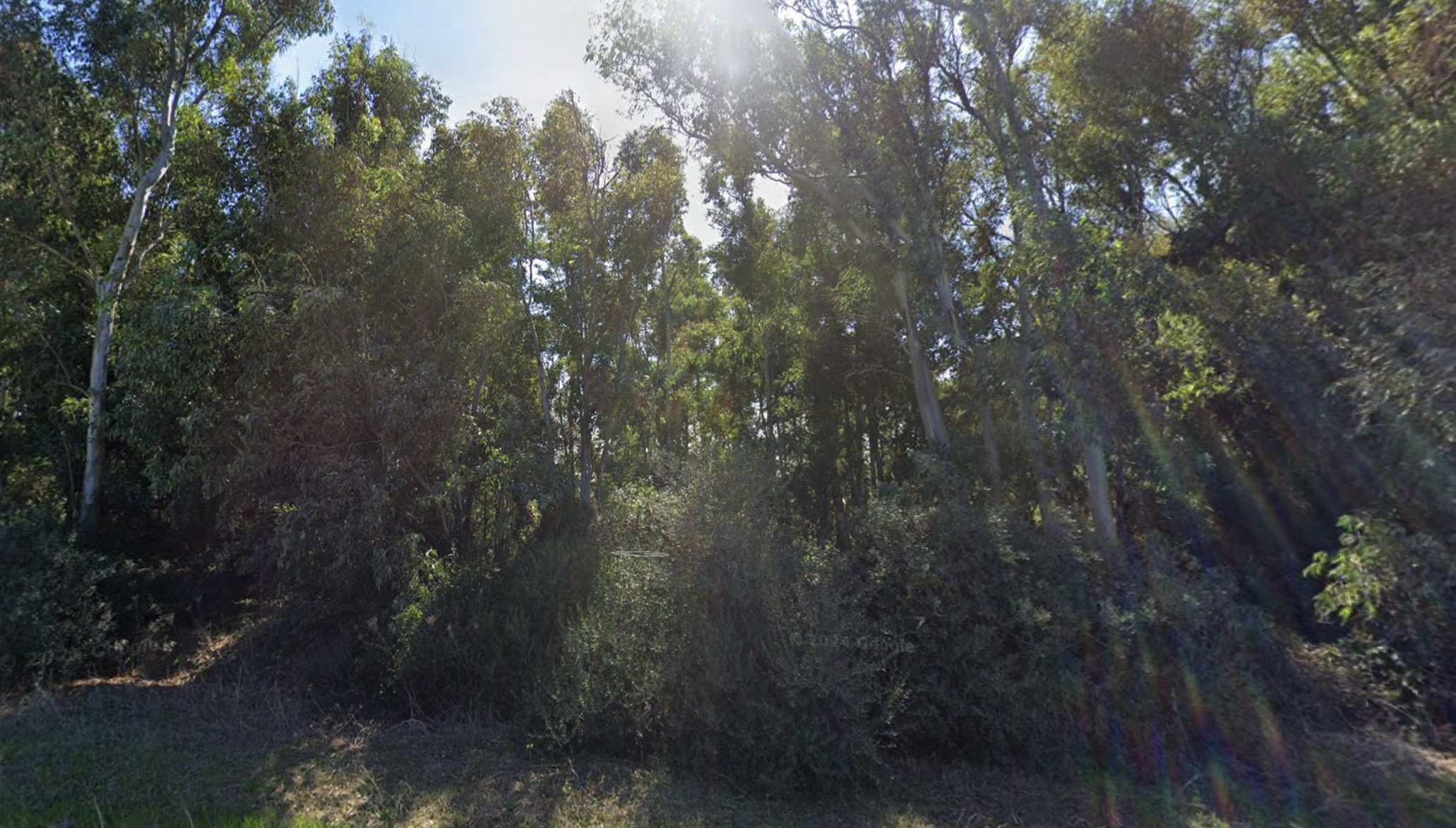}
     \caption{Real Dataset. We collected Google Street View images at a set of location where AGB measurements are provided by the Italian Arma dei Carabinieri~\cite{INFC}. Each row shows three images for different points (ID), having AGB measurements of $2.894$, $4.193$, $5.493$, which are on average predicted as $2.880$, $4.313$, $5.440$.}
    \label{fig:realdata}
\end{figure*}
\subsection{Dataset}
Our dataset is composed of a subset of the scenes in SPREAD, namely the Birch Forest and the Broadleaf Forest scene datasets. We discarded the Meadow Forest as it does not include instance segmentation maps. We also discarded the Redwood Forest since most of the images are a close-up view of a tree trunk captured close to the ground level. We also did not consider the Rainforest scene, as our focus is on temperate forests. Moreover, we found that in many assets of the Rainforest the DBH value is missing. We also did not consider urban scenes and plantations, since they are outside the scope of our work. In Equation~\ref{eq:agb} we set a density of $\rho=0.65\,g/cm^3$ for the Birch Forest, and a density of $\rho=0.55\,g/cm^3$ for the Broadleaf Forest~\cite{miles2009specific}.
We compute the AGB maps for all the images in the dataset, and split the data into $80\%$ for training and $20\%$ for testing, obtaining $5660$ training samples.
Figure~\ref{fig:histograms} shows the histograms of the AGB values in the single scene datasets and globally. 
\subsection{Methods}
We compare our architecture with suitable alternatives to demonstrate the effectiveness of our network design. 
We define a baseline system that outputs the median values of the AGB maps in the dataset. Our approach should do better than this simple prediction to demonstrate that the system has learned to interpret the RGB images.
Inspired by the performance of the Marigold depth estimation method~\cite{marigoldke2023}, we design a network for AGB map predictions that follow the same principle, namely we fine-tune the U-Net of a Latent Diffusion Model. We refer to this model as LDM in Table~\ref{table:evaluation}. 
We also considered a model composed of a DINO backbone and a UNet~\cite{unetronneberger2015}, where we fine-tune the DINO encoder using LoRA~\cite{hu2022lora} (DINO-LoRA-UNet). 
To investigate different options for the encoder, we replace the Swin Transformer (\textnormal{\texttt{swin\_large\_patch4\_window7\_224}}) with a DINOv2~\cite{oquab2023dinov2} encoder (Our (DINOv2)). 
Additionally, we tested an alternative approach where the model directly regresses the total AGB value instead of predicting a spatial density map (Our (no map)). In this setup, the extracted backbone features are passed through a processing block consisting of two linear layers separated by a ReLU activation function. 
In Our (depth) we report results for the case where we predict the depth maps during training.
\begin{figure*}
     \centering
    \includegraphics[width=0.28\linewidth]{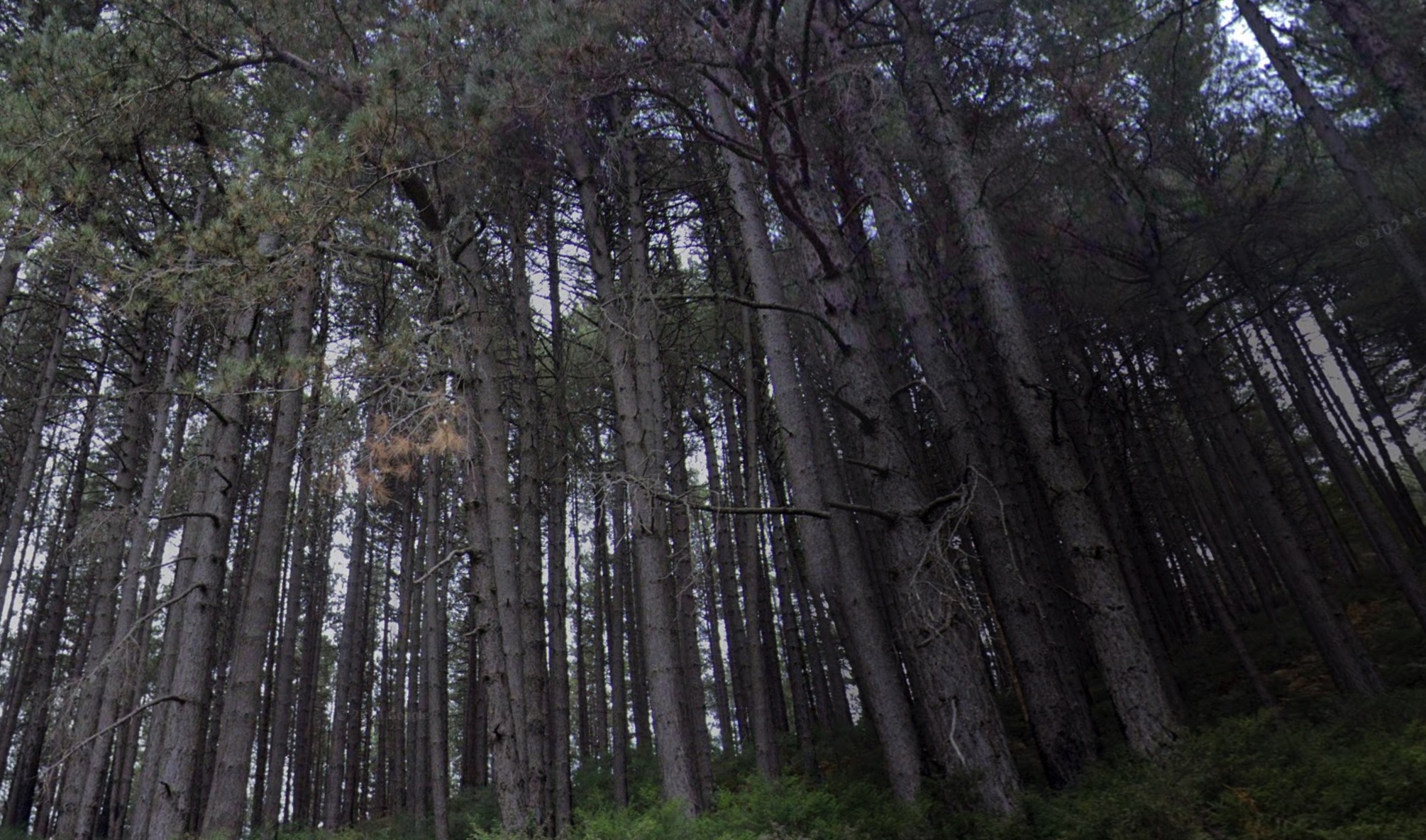}
    \includegraphics[width=0.28\linewidth]{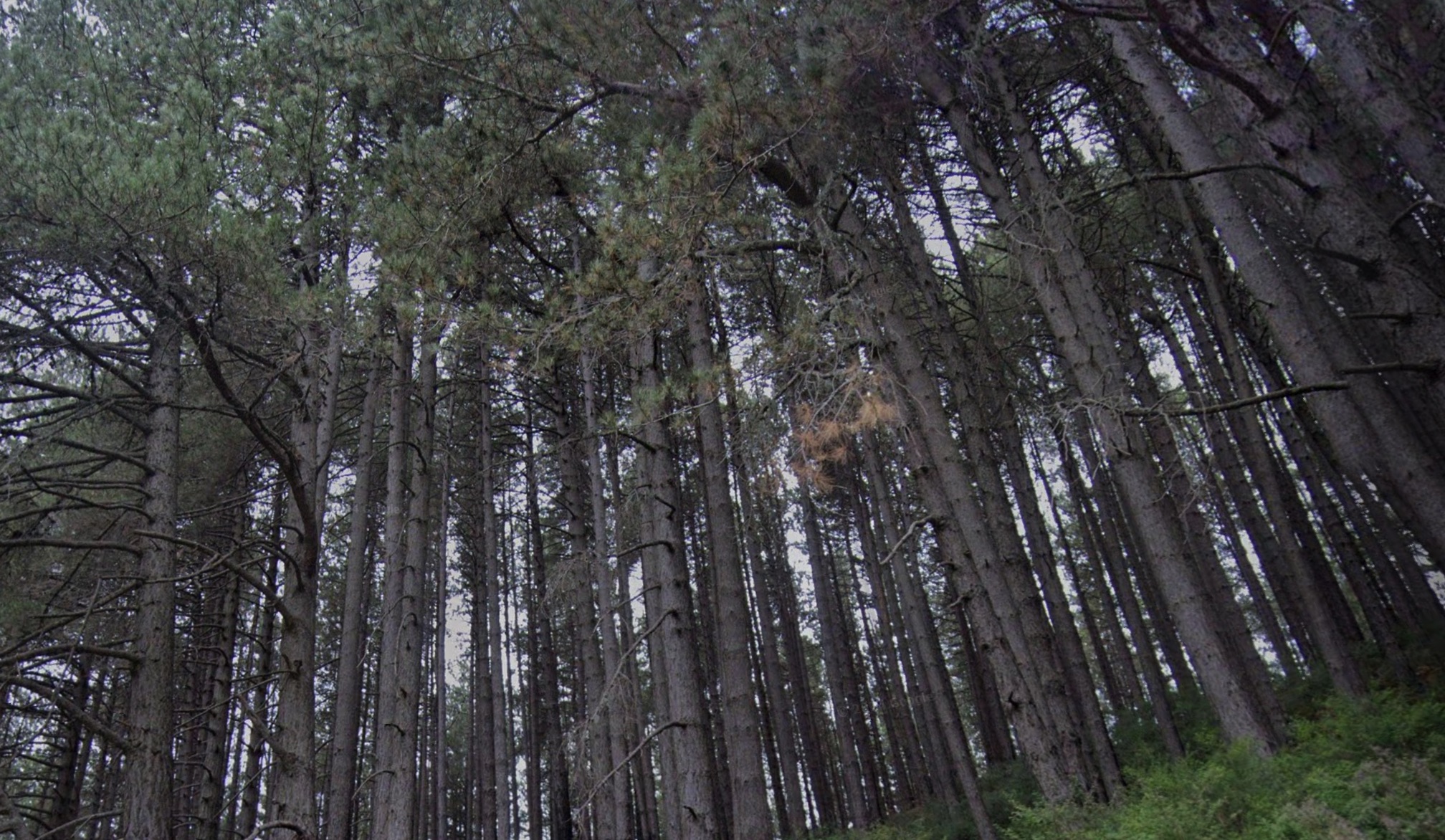}
    \includegraphics[width=0.28\linewidth]{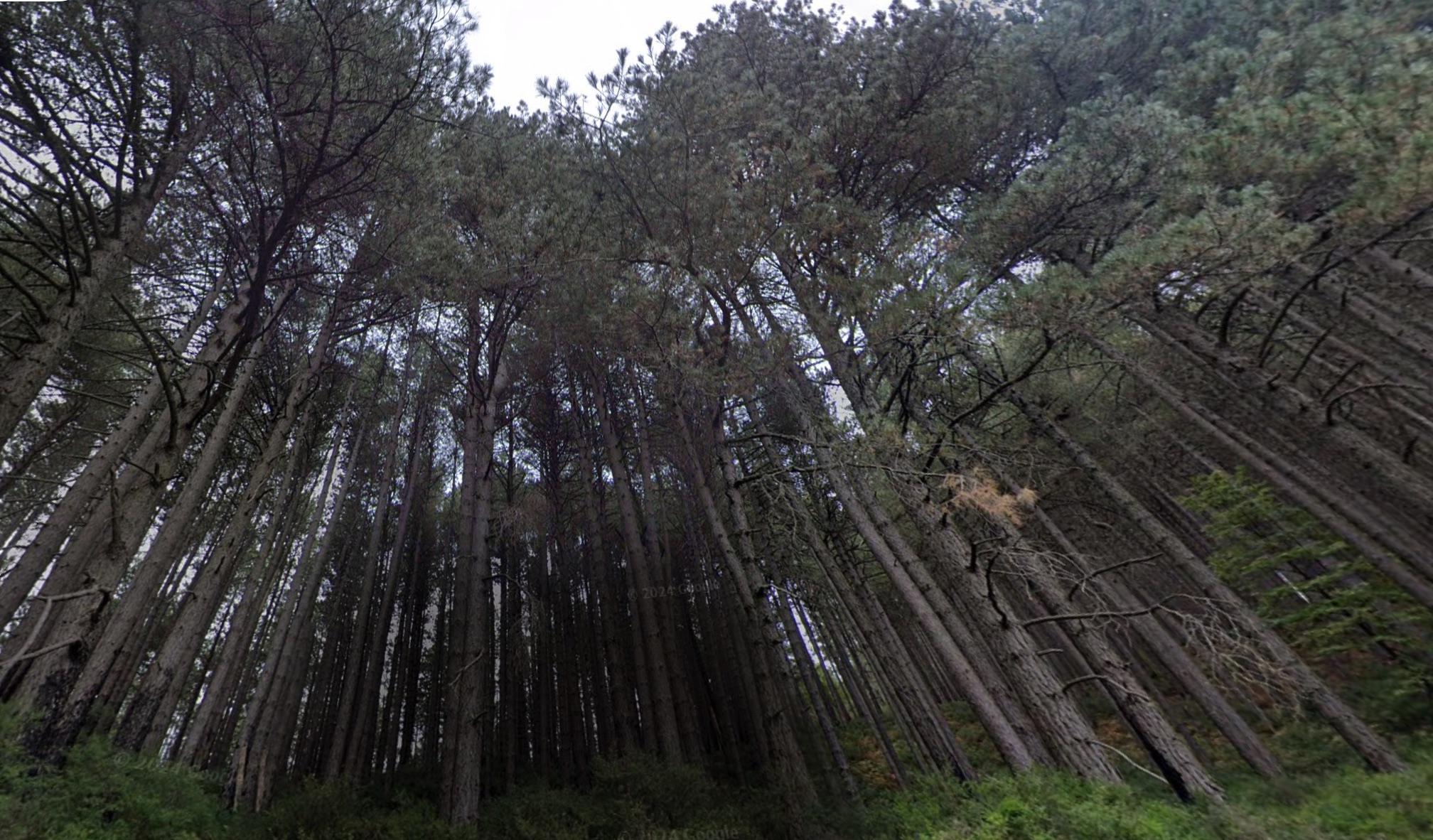}
   \centering
    \includegraphics[width=0.28\linewidth]{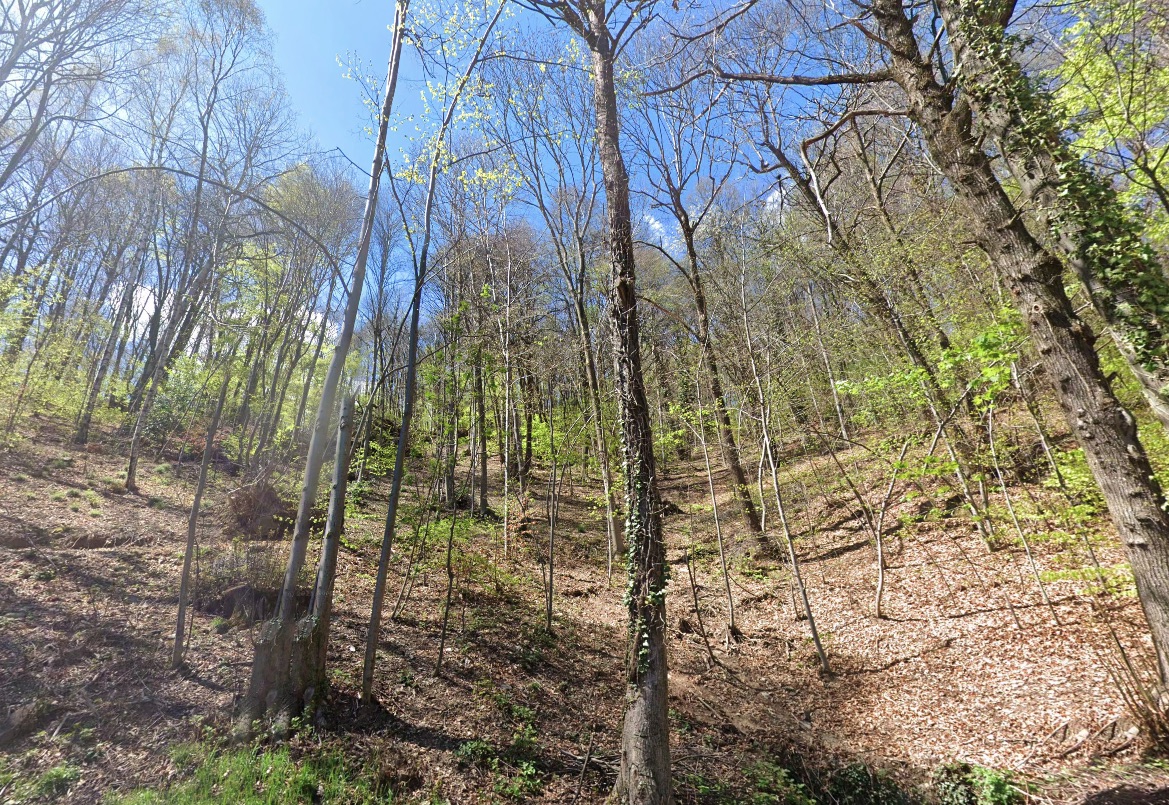}
    \includegraphics[width=0.28\linewidth]{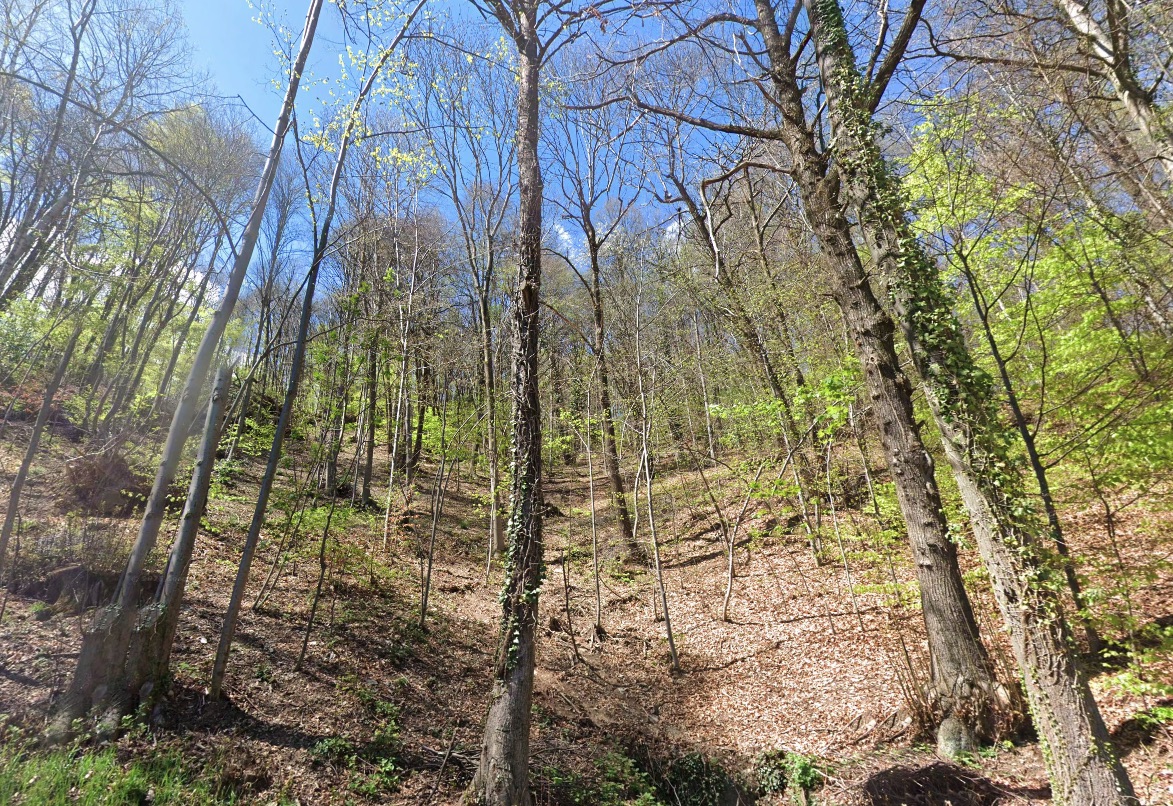}
    \includegraphics[width=0.28\linewidth]{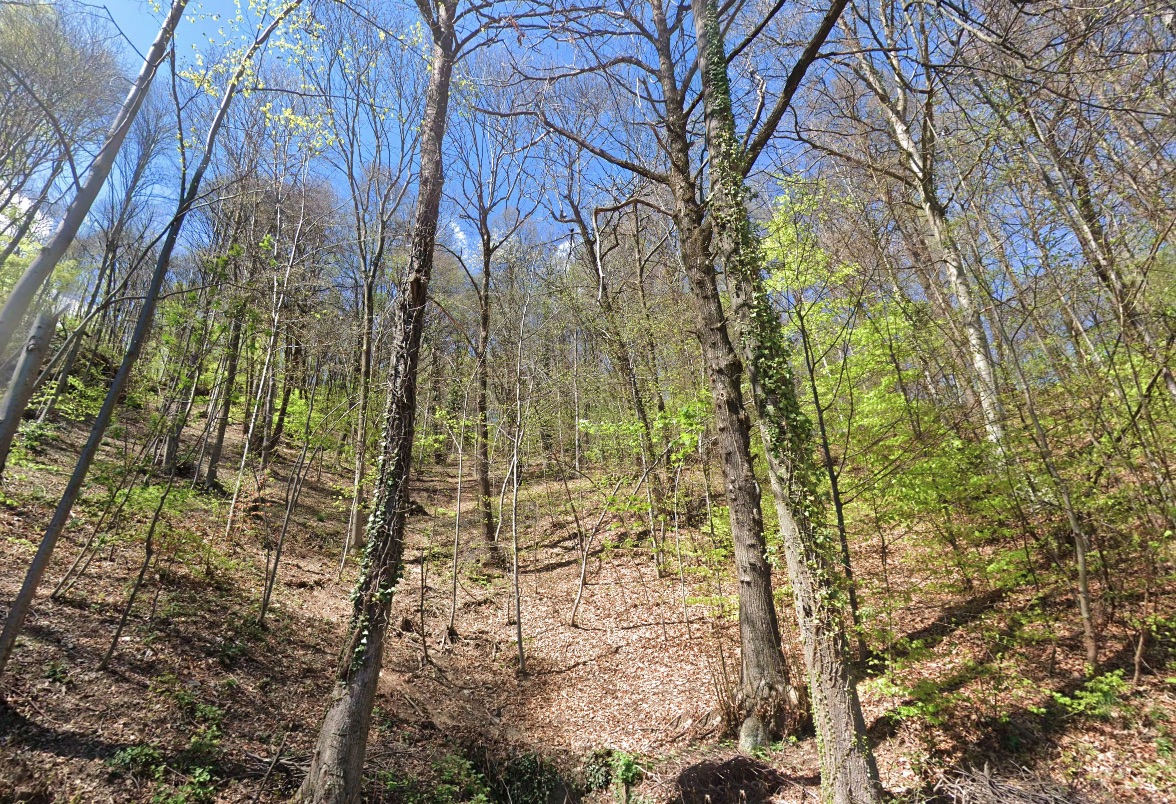}
     \centering
   \includegraphics[width=0.28\linewidth]{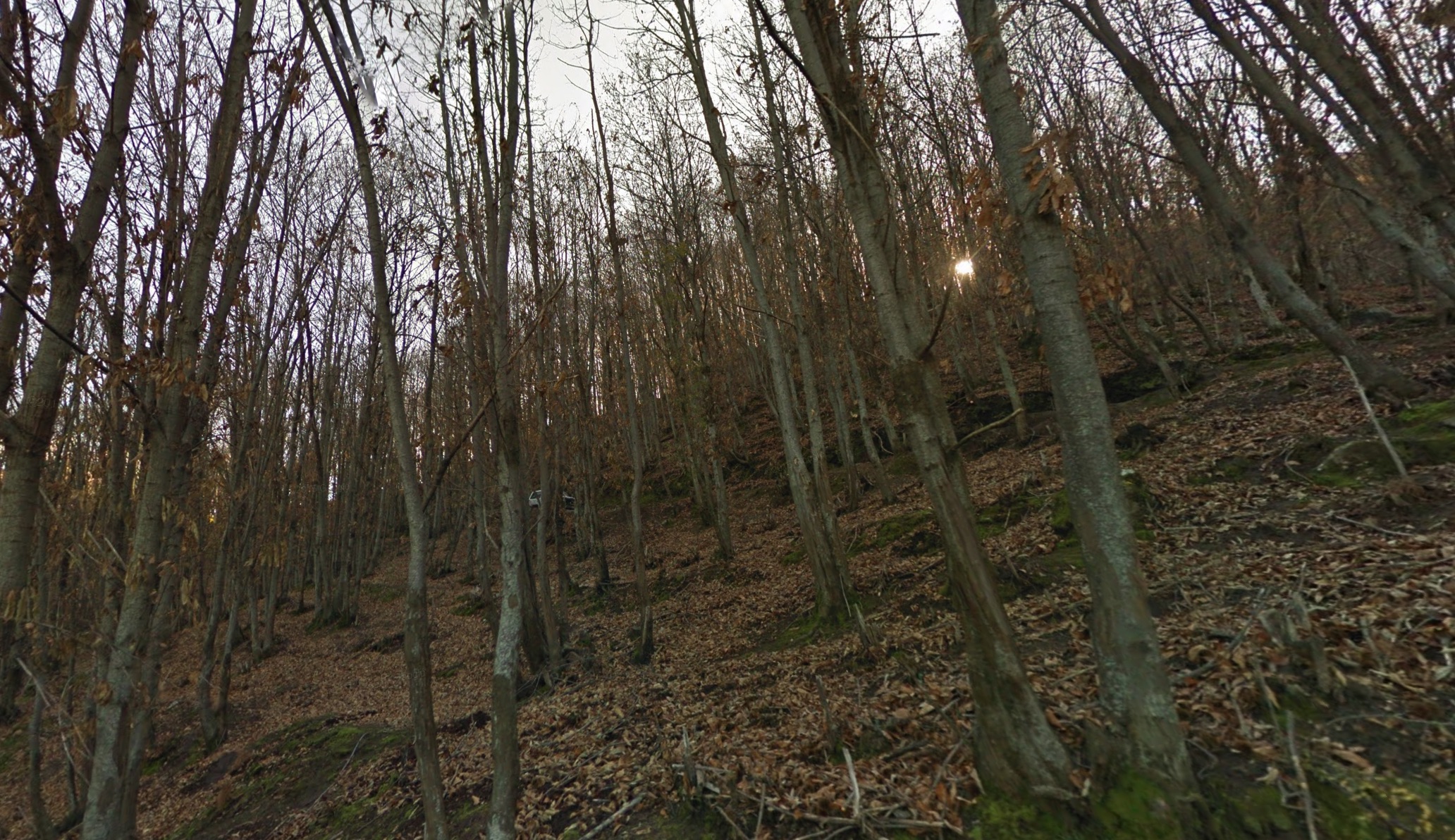}
   \includegraphics[width=0.28\linewidth]{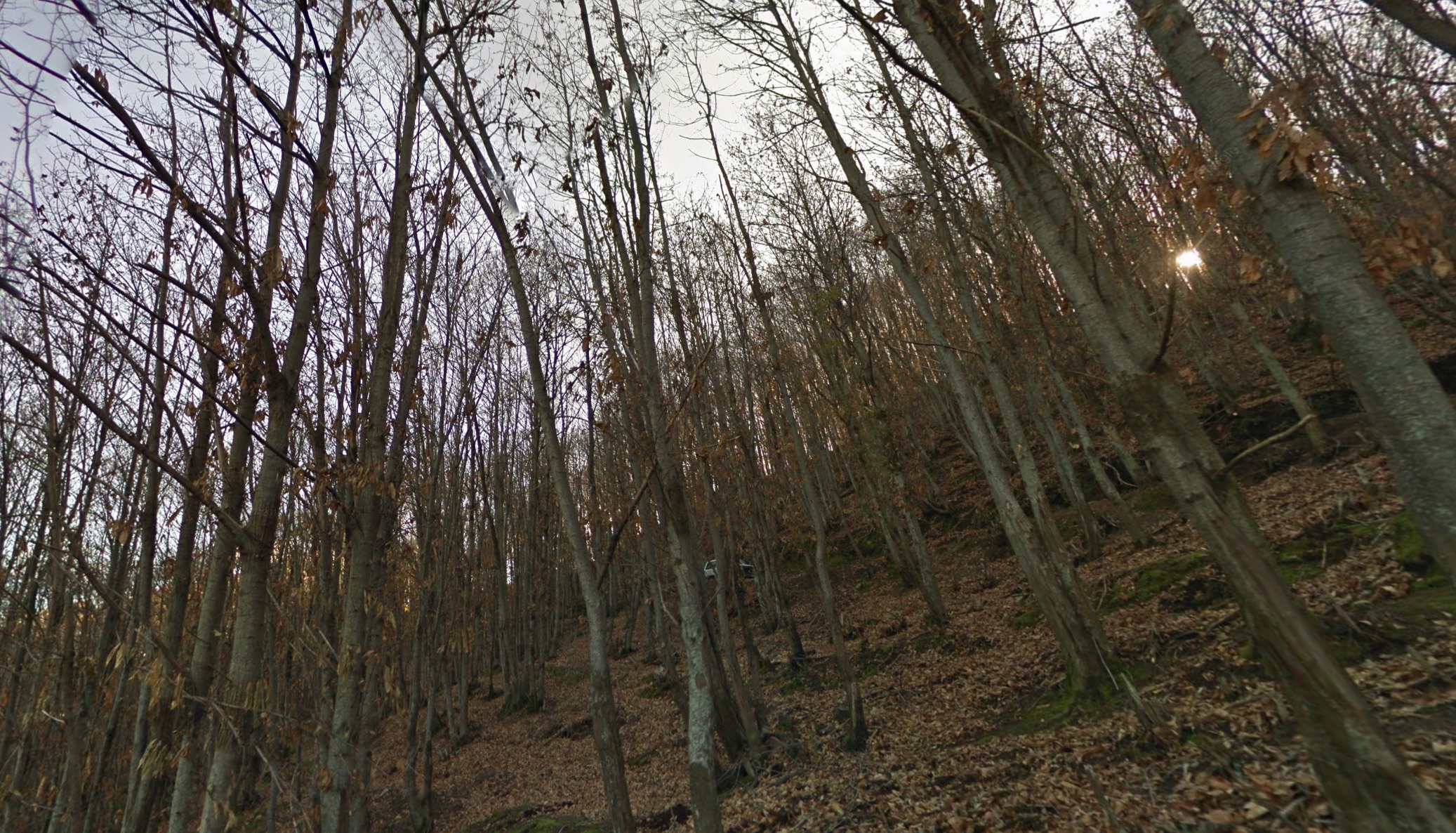}
   \includegraphics[width=0.28\linewidth]{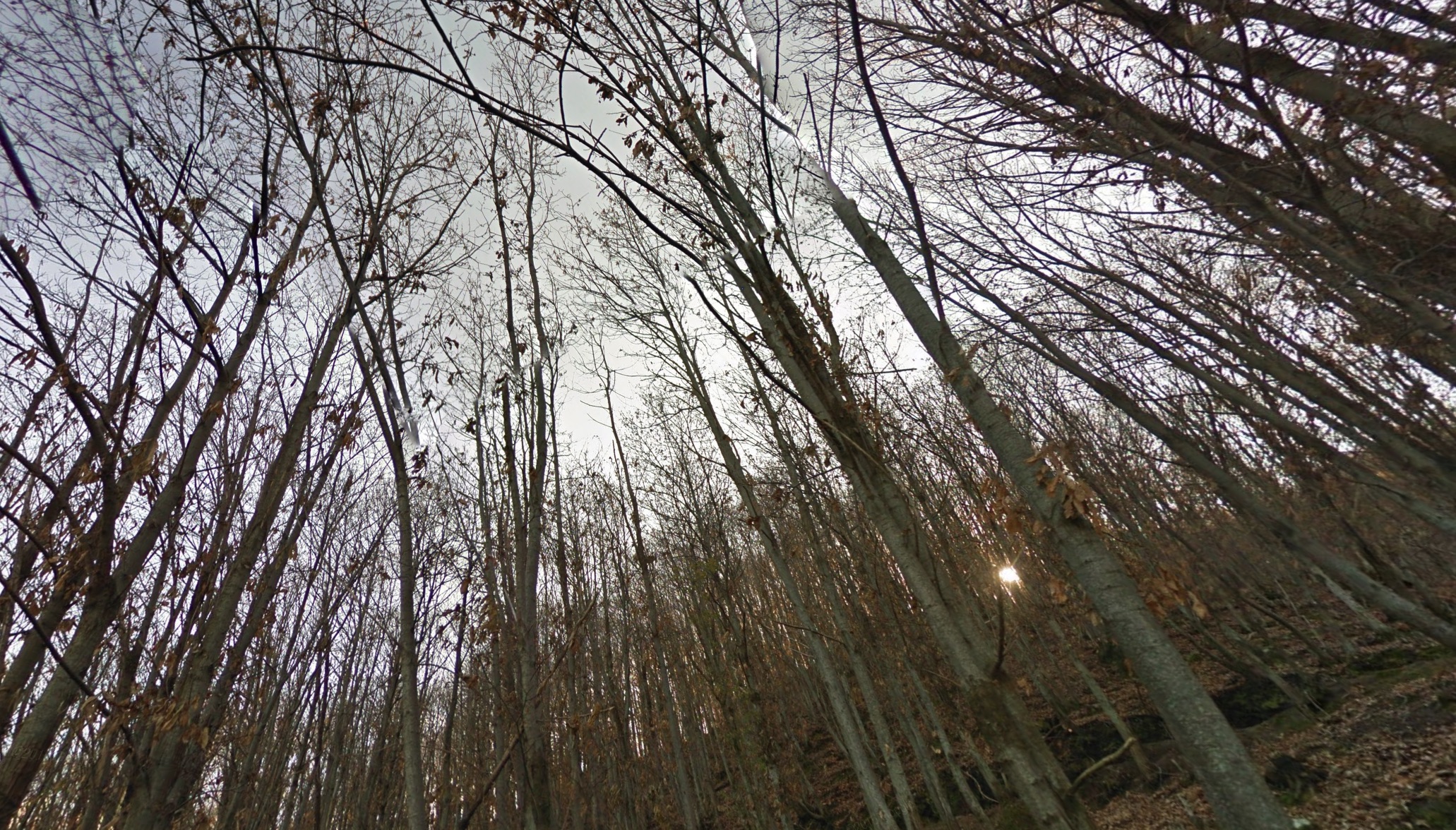}
     \caption{Challenging cases. We show here two cases with large prediction error. 
     The top row shows samples from a location with ground-truth AGB of $6.963$, with average prediction of $13.289$. 
     The middle row shows samples from a location with ground-truth AGB of $18.987$, with average prediction of $3.640$. 
     The bottom row shows samples with ground-truth AGB of $14.356$, with average prediction of $7.424$.} 
    \label{fig:realdata2}
\end{figure*}

\subsection{Training Details}
We train the networks with the full-size images, and precomputed AGB density maps. Note that applying random crops to the images and the AGB density maps would result in incorrect density maps. 
Therefore, we augment the training set with random image flip along the vertical axis, which preserves the correctness of the data.
We train all networks for 40 epochs, after which the training loss stabilizes.
The LDM model is an exception. Due to its higher complexity, we limited the training to $4$ epochs, after which we observed an increase in the training loss. 

\subsection{Analysis} 
\begin{figure*}[!htbp]
    \centering
    \includegraphics[width=\linewidth]{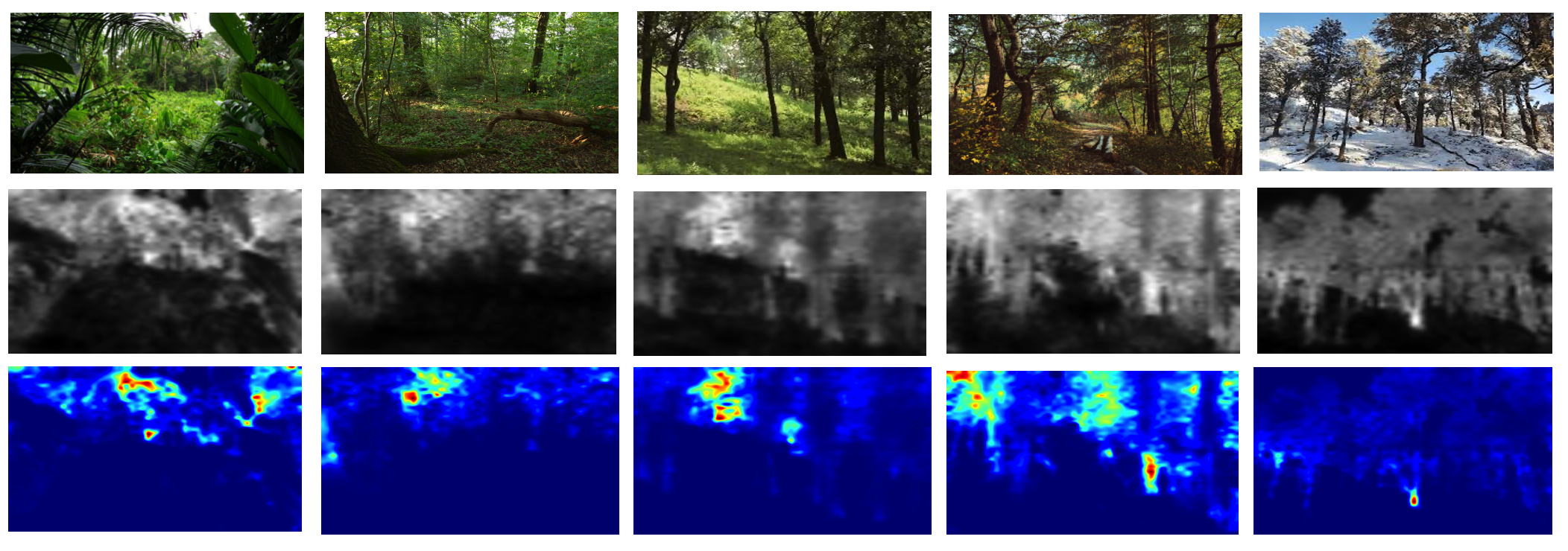}
     \centering
    \includegraphics[width=\linewidth]{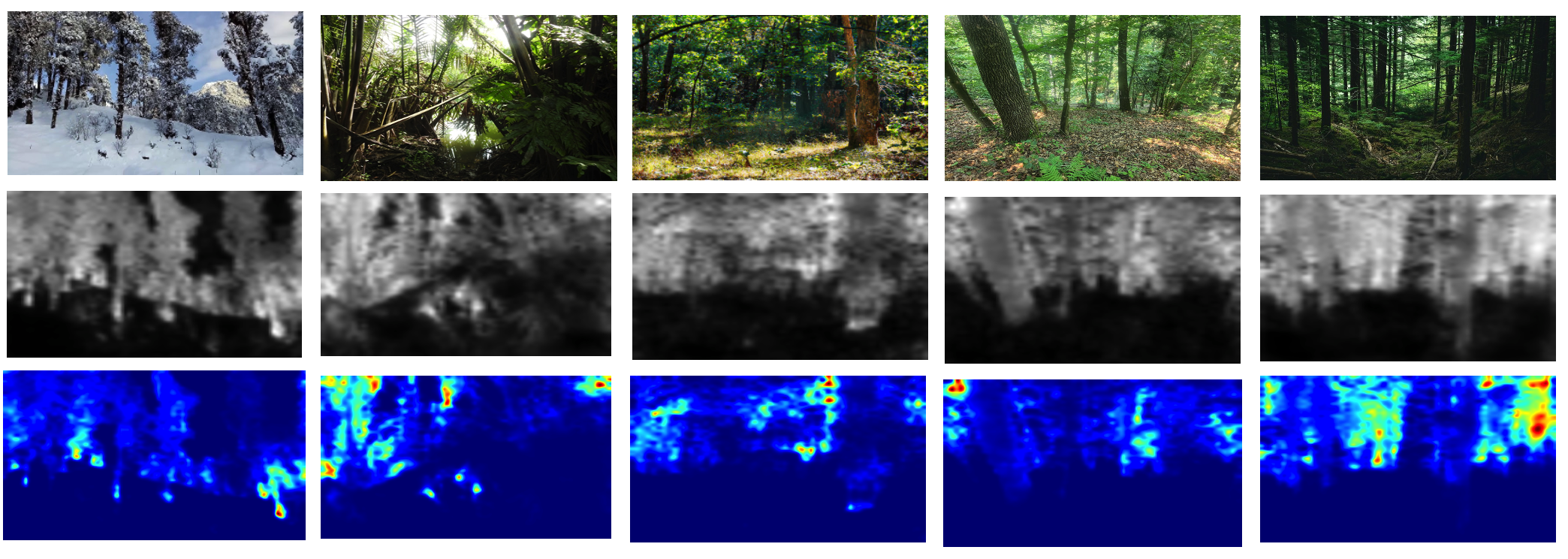}
    \centering
    \includegraphics[width=\linewidth]{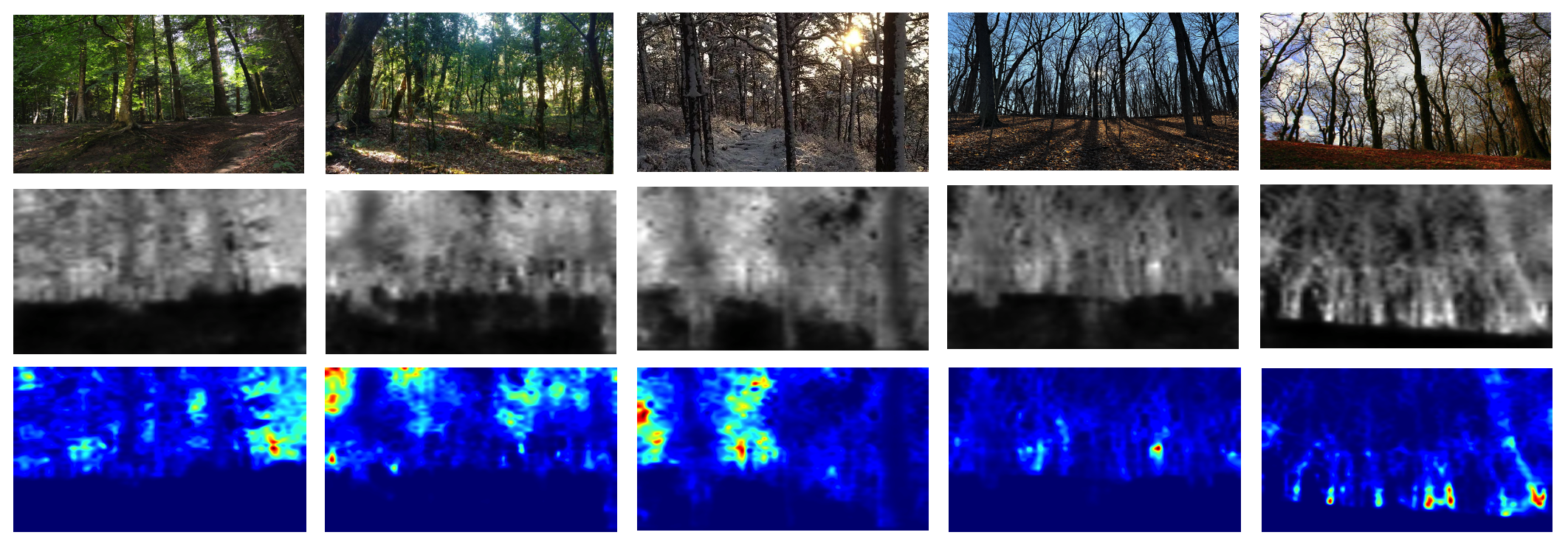}
     \caption{Application to real images. We apply our model to real images, and show them ordered by increasing total AGB, that we compute by integrating the predicted AGB density maps. Images are visualized in three groups, and in each group we show in the first row the RGB image, below the AGB map, and the AGB map with colormap base visualization. Note that the AGB map images correspond to AGB density maps normalized to their maximum value for visualization, and therefore inter-map comparisons are not meaningful. Moreover, note that the grayscale AGB maps are displayed after non-linear enhancement to better visualize the detected forest structure. Images have been collected from the Web and represent a variety of type of forests in different weather conditions. It is interesting to observe that the images at the top row (lower AGB) contain few small trees, while the images at the bottom (higher AGB) contain many tall trees.}
    \label{fig:real}
\end{figure*}

We report our results in terms of absolute error, as this metric is more robust to outliers.
Outliers may occur when the assumption of tree presence, used to approximate the plot area, is not satisfied, leading to abnormally large AGB values in certain maps.
We observe from Table~\ref{table:evaluation} that the proposed model (Our) outperforms the alternatives we have investigated.
As a further analysis, we train separate models for each scene dataset (Our (Birch) and Our (Broadleaf) and compute the errors within each scene test set. 
Results are reported in Table~\ref{table:per_scene_models_evaluation} and in Table~\ref{table:per_scene_models_evaluation_no_map} for the case without AGB map prediction.
In Table~\ref{table:per_scene_models_evaluation} the performance of the model trained with all the data is in general better than those for single datasets, illustrating that the model does not overfit to single datasets and instead benefits from the larger number of training samples.
In Table~\ref{table:per_scene_models_evaluation_no_map}, where we show a similar experiment for models trained without dense AGB map prediction, we do not observe a similar behavior, as the model trained exclusively on the Broadleaf Forest data (Our (no map, Broadleaf)) performs better than the global model (Our (no map)). Additionally, observe again, in Table~\ref{table:per_scene_models_evaluation}, the error difference when using the model trained on the Broadleaf Forest training set (Our (no map, Broadleaf)) on the Birch Forest test set (top section) with respect to the model trained on Birch Forest data (Our (no map, Birch)): $4.29$ compared with $1.58$ for the average error. In particular, compare these results with Table~\ref{table:per_scene_models_evaluation}, where the average error is $2.42$ instead of $4.29$. This indicates that predicting the density AGB map leads to a more robust regression model.
In Figure~\ref{fig:real}, we apply our method to a set of images downloaded from the Web, displaying them in ascending order of AGB predictions. It is interesting to observe that images with fewer and shorter trees are correctly ranked before those containing a larger number of taller trees. This demonstrates that the model we have trained exclusively on synthetic data can perform reasonably on real data. 

\subsection{Real Dataset}
\begin{figure}[t]
    \centering
    \includegraphics[width=\linewidth]{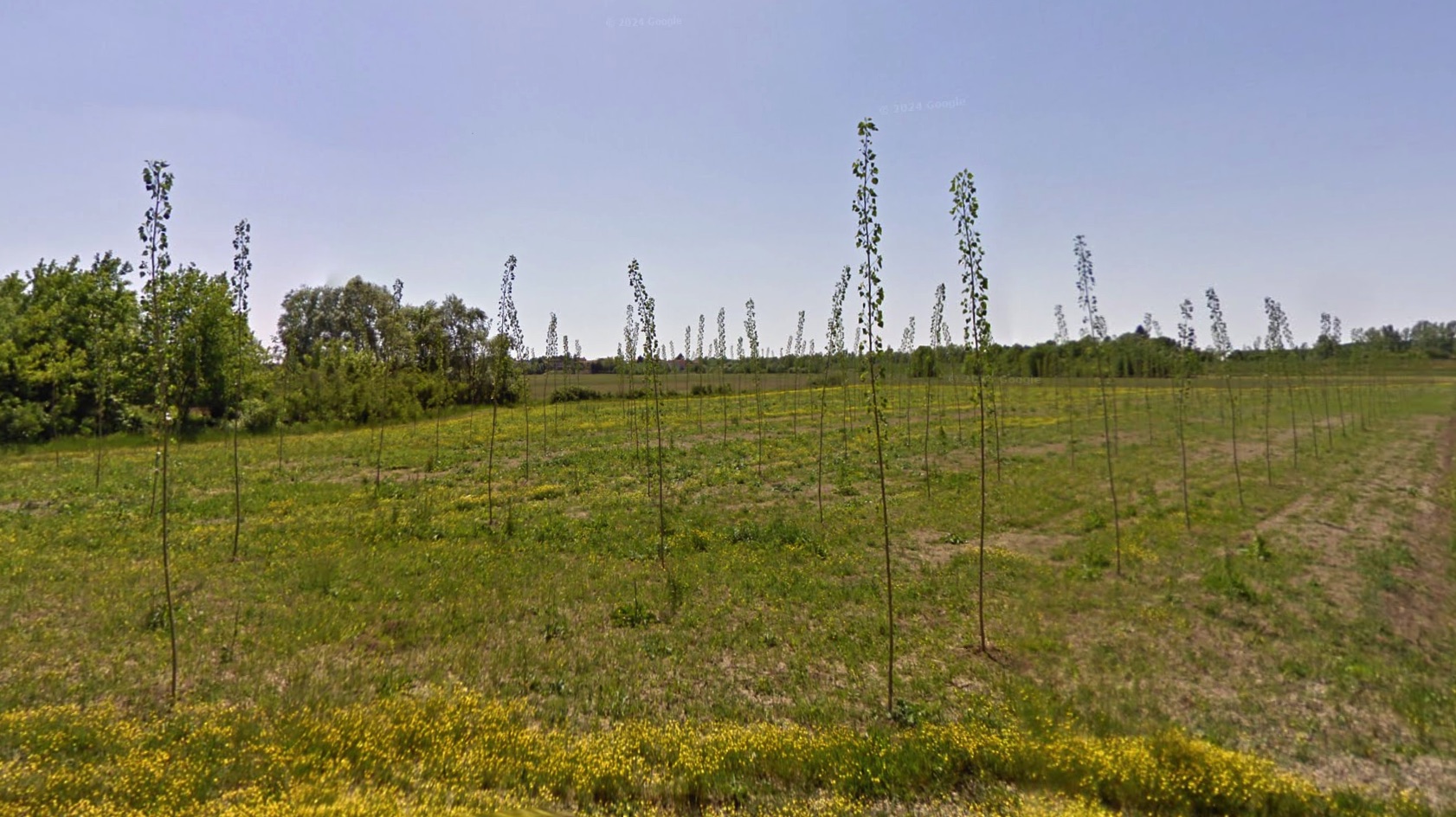}
     \caption{Real dataset. An example of discrepancy between the AGB ground-truth and the Street View images. Here the reported ground-truth AGB value is $5.674$ for a poplar plantation. Clearly the image refers to the early stages of the plantation, with a lower biomass.}
    \label{fig:poplar}
\end{figure}

\begin{figure}[t]
    \centering
    \includegraphics[width=\linewidth]{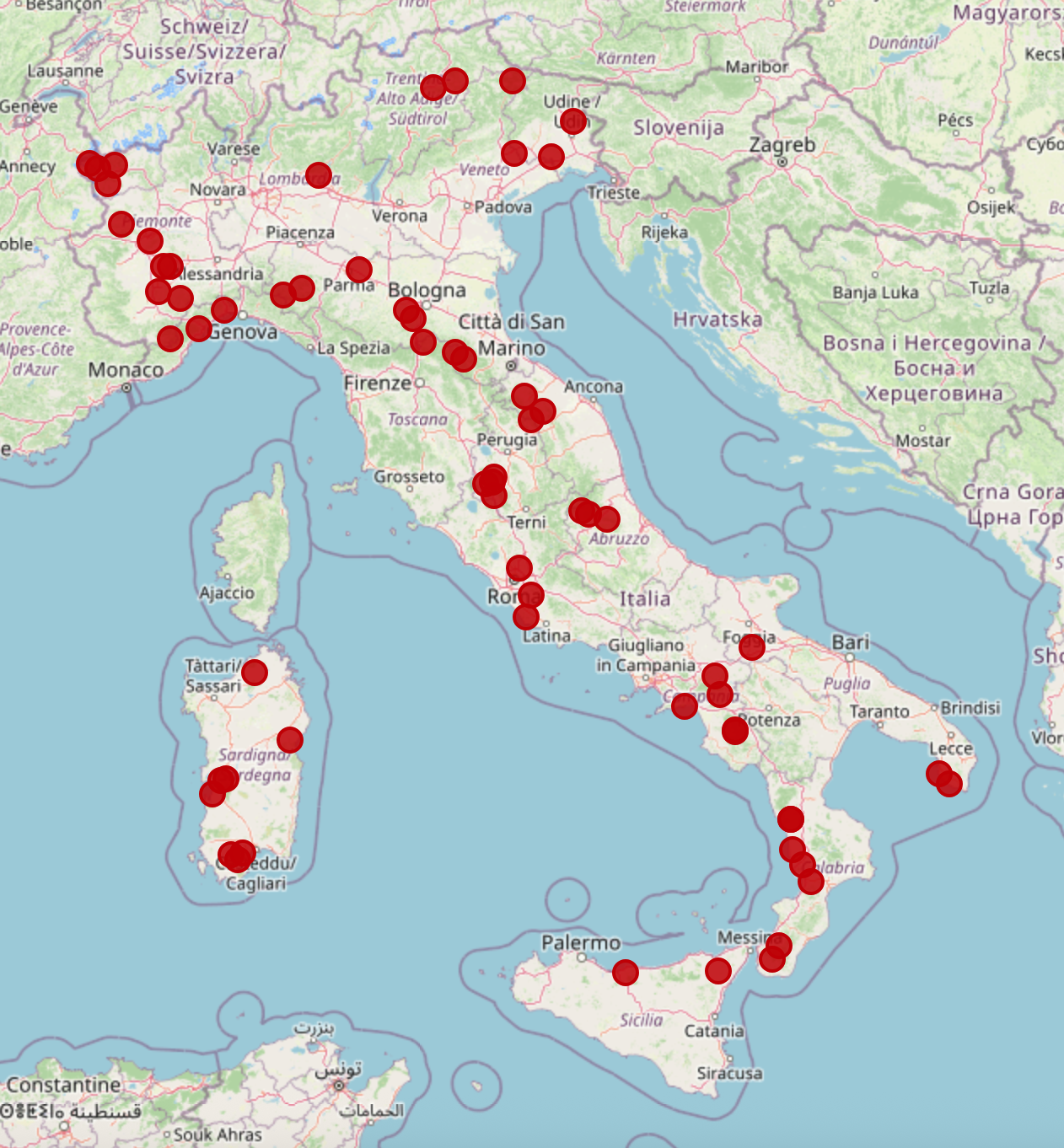}
     \caption{Locations where we have collected Google Street View images to pair with AGB values provided by the Italian National Forests Inventory~\cite{INFC}.}
    \label{fig:italy}
    \end{figure}

Quantitative experiments have been performed on held-out data from the SPREAD dataset. To our knowledge, datasets with ground-level images and corresponding AGB do not exist. In order to attempt a quantitative evaluation on real data, we did the following. We collected AGB values on a set of points in Italy from the Italian National Forests Inventory~\cite{INFC}. 
The inventory data includes a set of information, among which the \textit{dry weight of total above-ground biomass of living trees, per hectare} and \textit{dry weight of shrubs, per hectare}, for different heights. We sum this data and convert to per square-meter units to define the AGB ground-truth. The data also includes a set of codes for different vegetation types. We filtered the data for vegetation types of \textit{beech, oaks, chestnut, hornbeam/hop-hornbeam, hygrophilous, other deciduous broadleaves, holm/cork oak, poplar and other broadleaf plantations}.
. 
We then created a clickable map of Italy, and used this map to access Google Street View data, where available in proximity of the inventory point, excluding points with AGB that exceeds $20\,kg/m^2$, as this aligns with the training distribution. For each site, we collected about $3$ images (see Fig.~\ref{fig:realdata} for a few examples). We then processed the images through our network to estimate the AGB values. We have $67$ points for a total of $207$ samples (see Fig.\ref{fig:italy}). It is worth observing that while creating this dataset, we noticed that for some points the AGB value reported was higher than expected (see Fig.\ref{fig:poplar}). This can be due to the fact that there is a time discrepancy between the time the AGB data was collected, and the time the Street View images were captured. Street View images can also exhibit significant fisheye distortion.
Table~\ref{table:total_agb_errors} reports prediction errors.
We also computed the Spearman’s rank correlation to evaluate the ability of the method to assign lower estimates to low AGB cases and higher estimates to large AGB cases. 
We believe this analysis partially addresses the time mismatch between the images and the ground truth AGB measurements.
The correlation was weak ($\rho$ = 0.16), indicating limited ranking ability. However, after pruning the worst 20\% of cases, the correlation increased substantially ($\rho$ = 0.43), suggesting that the estimator has a moderate ability to preserve the ranking of AGB values once extreme outliers are removed.

\begin{table}[t]
\centering
{
\setlength{\tabcolsep}{6pt} 
\renewcommand{\arraystretch}{1.3} 
\begin{tabular}{lcc}
\toprule
\multicolumn{1}{c}{} & \multicolumn{2}{c}{Prediction Error ($kg/m^2$)} \\
\cmidrule(lr){2-3} 
Method   &  Mean &  Median \\
\midrule
\midrule
LDM &3.85 &1.49 \\ 
DINO-LoRA-UNet  &5.39 &3.58  \\ 
 
\midrule
Our (DINOv2) &6.87 &4.71\\ 
Our (no map) &3.21 &1.24\\ %
Our (depth)  &3.15 &1.25\\ 
Our &\textbf{3.00} &\textbf{1.22}\\ %
\midrule
Baseline &6.09 &3.70\\ 
\bottomrule
\end{tabular}
}
\caption{Absolute error statistics. We compare our proposed architecture with a set of alternative solutions for dense prediction tasks (see text). Baseline refers to a method that predicts the median AGB in the training set. Our solution achieves a median prediction error of $1.22\,kg/m^2$ ($22\%$). }
 ]
\label{table:evaluation}
\end{table}

\begin{table}[t]
\centering

{
\setlength{\tabcolsep}{4pt} 
\renewcommand{\arraystretch}{1.3} 
\begin{tabular}{lcccc}
\toprule
\multicolumn{1}{c}{} & \multicolumn{4}{c}{Prediction Error ($kg/m^2$)} \\
\multicolumn{1}{c}{} & \multicolumn{2}{c}{Birch} & \multicolumn{2}{c}{Broadleaf} \\
\cmidrule(lr){2-3} \cmidrule(lr){4-5}
Method & Mean & Median & Mean & Median \\ 
\midrule
\midrule
Our (Birch)     & 1.55 & \textbf{0.88} & 6.33 & 2.59 \\ 
Our (Broadleaf)  & 2.42 & 1.76 & 5.56 & 2.02 \\ 
Our            & \textbf{1.51} & 0.89 & \textbf{4.74} & \textbf{1.80} \\ 
\midrule
Baseline                & 3.51 & 2.49 & 8.43 & 5.82 \\ 
\bottomrule
\end{tabular}
}
\caption{Absolute error statistics. 
We compare models trained on scene specific datasets. Our (Birch Forest) is trained only on images from the Birch Forest scene and Our (Broadleaf Forest) is trained only on images from the Broadleaf Forest scene. We test both models on separate per-scene test sets. Note that performance for crossed scenes are comparable, and lower than the baseline.}
\label{table:per_scene_models_evaluation}
\end{table}

\begin{table}[t]
\centering
{
\setlength{\tabcolsep}{4pt} 
\renewcommand{\arraystretch}{1.3} 
\begin{tabular}{lcccc}
\toprule
\multicolumn{1}{c}{} & \multicolumn{4}{c}{Prediction Error ($kg/m^2$)} \\
\multicolumn{1}{c}{} & \multicolumn{2}{c}{Birch} & \multicolumn{2}{c}{Broadleaf} \\
\cmidrule(lr){2-3} \cmidrule(lr){4-5}
Method & Mean & Median & Mean & Median \\ 
\midrule
\midrule
Our (no map, Birch)     & 1.58 & 1.01 & 6.07 & 2.30 \\ %
Our (no map, Broadleaf)  & 4.29 & 2.94 & \textbf{4.86} & 1.90 \\ %
Our (no map)            & \textbf{1.52} & \textbf{0.92} & 5.21 & \textbf{1.85} \\ 
\midrule
Baseline                & 3.51 & 2.49 & 8.43 & 5.82 \\ 
\bottomrule
\end{tabular}
}
\caption{Absolute error statistics. We compare models trained on scene specific datasets to regress directly the overall AGB (see text). Note the large error difference when applying the model trained on the Broadleaf Forest (Our (no map, Broadleaf)) to the Birch Forest with respect to the model trained on Birch Forest data (Our (no map, Birch)). In particular, compare these results with Table~\ref{table:per_scene_models_evaluation}, where the average error is $2.42$ instead of $4.29$. This suggests that predicting the AGB density map results in a more robust regressor.}
\label{table:per_scene_models_evaluation_no_map}
\end{table}

\begin{table}[t]
\centering
{
\setlength{\tabcolsep}{4pt} 
\renewcommand{\arraystretch}{1.3} 
\begin{tabular}{lcccccc}
\toprule
\multicolumn{1}{c}{} & \multicolumn{6}{c}{Prediction Error ($kg/m^2$)} \\
\multicolumn{1}{c}{} & \multicolumn{3}{c}{Per-ID} & \multicolumn{3}{c}{All-samples}\\
\cmidrule(lr){2-4} \cmidrule(lr){5-7} 
 Method & Mean & Median & Std & Mean & Median & Std \\
\midrule
\midrule
Our & 3.14 & 3.11 & 0.74 & 3.10 &1.94 & 3.15  \\
\bottomrule
\end{tabular}
}
\caption{Absolute error statistics on real data. Per-ID: computed by averaging over samples for each point; All-samples: computed directly across all images; STD per-ID refers to the standard deviation for each ID. We achieve an error of $1.94\,kg/m^2$ ($44\%$).} 
\label{table:total_agb_errors}
\end{table}

\section{Conclusion}
\vspace{-2mm}
We address the challenge of predicting Aboveground Biomass (AGB) from a single RGB image. Our approach introduces a novel two-dimensional representation, the AGB density map, which encodes per-pixel AGB density for individual trees within the captured area. To develop this method, we leverage a highly realistic synthetic 3D dataset that provides RGB images together with tree geometric parameters, enabling the computation of corresponding biomass values through allometric equations.
We train a Swin Transformer–based regressor to predict AGB density maps directly from RGB inputs. Quantitative evaluation on a held-out portion of the dataset shows that our method can predict scene-level AGB with a median error of $1.22\,kg/m^2$. We also perform both qualitative and quantitative evaluations on real-world images.
Specifically, we apply our method to a set of web-sourced images, order them by increasing predicted AGB, and show that scenes with low predicted AGB contain fewer trees than those with high predicted AGB. While quantitative validation is challenging due to the lack of ground-truth data, we construct a real-image dataset by collecting Google Street View images at sites where measured AGB values are available from the Italian National Forest Inventory~\cite{INFC}. On this dataset, our method achieves a median error of $1.94\,kg/m^2$.
Although this dataset is affected by certain limitations, such as location inaccuracies (Street View images may not perfectly match inventory sites), distortions in the RGB images (due to differences between Street View cameras and the one assumed in the SPREAD dataset), and temporal mismatches between image acquisition and inventory measurements, we find that our method yields reasonable results. These findings confirm the suitability of our novel approach for estimating AGB from a single RGB image.
The field of computer vision is rapidly progressing toward the development of world models with metric accuracy. We believe that our approach will benefit from these advances and encourage the creation of real-world datasets to evaluate the effectiveness of AI-based methods for ecosystem monitoring.

\noindent\textbf{Acknowledgements}.  
SZ is funded by NRRP, Miss. 4 Comp. 2 Inv. 1.4 - Call No. 3138 16/12/21, rect. by Decree n.3175 18/12/21  of MUR funded by NextGenerationEU; Award N.: Proj. code CN00000033, Conc. Decree No. 1034 17/06/22 CUP B83C22002930006, title: National Biodiversity Future Center - NBFC.

{
   \bibliographystyle{eg-alpha-doi}
   \bibliography{egbib}
}
\end{document}